\documentclass{article}

\usepackage{amsmath,amsfonts,bm}

\def\eqref#1{equation~\ref{#1}}

\def\1{\bm{1}}

\DeclareMathAlphabet{\mathsfit}{\encodingdefault}{\sfdefault}{m}{sl}
\SetMathAlphabet{\mathsfit}{bold}{\encodingdefault}{\sfdefault}{bx}{n}

\usepackage{microtype}
\usepackage{graphicx}
\usepackage{subfigure}
\usepackage{array}

\newcolumntype{H}{>{\setbox0=\hbox\bgroup}c<{\egroup}@{}}
\usepackage{lipsum}  
\usepackage{xcolor,soul}

\usepackage{booktabs} 
\usepackage{multirow}  

\usepackage{amssymb}
\usepackage{pifont}
\newcommand{\cmark}{\ding{51}}%
\newcommand{\xmark}{\ding{55}}%
\usepackage{hyperref}

\usepackage{algorithmic}

\usepackage{soul,xcolor}

\usepackage{todonotes}
\definecolor{MutedRed}{RGB}{255, 0, 0} 
\definecolor{MutedBlue}{RGB}{0, 0, 255} 
\definecolor{Black}{RGB}{0, 0, 0} 
\newcommand{\bR}{\color{MutedRed}\boldmath}
\newcommand{\bK}{\color{Black}\boldmath}
\newcommand{\bB}{\color{MutedBlue}}

\newcommand{\GN}{NASFLAT}
\definecolor{moh_colour}{RGB}{255, 204, 205}

\newcommand{\comment}[1]{}

\usepackage[accepted]{mlsys2024}

\mlsystitlerunning{On Latency Predictors for Neural Architecture Search}

\begin{document}

\twocolumn[
\mlsystitle{On Latency Predictors for Neural Architecture Search}



\begin{mlsysauthorlist}
\mlsysauthor{Yash Akhauri}{to}
\mlsysauthor{Mohamed S. Abdelfattah}{to}
\end{mlsysauthorlist}

\mlsysaffiliation{to}{Cornell University}

\mlsyscorrespondingauthor{Yash Akhauri}{ya255@cornell.edu}

\mlsyskeywords{Machine Learning, MLSys}

\vskip 0.3in

\begin{abstract}
Efficient deployment of neural networks (NN) requires the co-optimization of accuracy and latency. 
For example, hardware-aware neural architecture search has been used to automatically find NN architectures that satisfy a latency constraint on a specific hardware device. 
Central to these search algorithms is a prediction model that is designed to provide a hardware latency estimate for a candidate NN architecture. 
Recent research has shown that the sample efficiency of these predictive models can be greatly improved through pre-training on some \textit{training} devices with many samples, and then transferring the predictor on the \textit{test} (target) device.
Transfer learning and meta-learning methods have been used for this, but often exhibit significant performance variability.
Additionally, the evaluation of existing latency predictors has been largely done on hand-crafted training/test device sets, making it difficult to ascertain design features that compose a robust and general latency predictor. 
To address these issues, we introduce a comprehensive suite of latency prediction tasks obtained in a principled way through automated partitioning of hardware device sets.
We then design a general latency predictor to comprehensively study (1) the predictor architecture, (2) NN sample selection methods, (3) hardware device representations, and (4) NN operation encoding schemes.
Building on conclusions from our study, we present an end-to-end latency predictor training strategy that outperforms existing methods on 11 out of 12 difficult latency prediction tasks, improving latency prediction by 22.5\% on average, and up to to 87.6\% on the hardest tasks. Focusing on latency prediction, our HW-Aware NAS reports a $5.8\times$ speedup in wall-clock time.
Our code is available on \href{https://github.com/abdelfattah-lab/nasflat_latency}{https://github.com/abdelfattah-lab/nasflat\_latency}.
\end{abstract}
]


\printAffiliationsAndNotice{}  

\section{Introduction}

With recent advancements in deep learning (DL), neural networks (NN) have become ubiquitous, serving a wide array of tasks in different deployment scenarios. 
With this ubiquity, there has been a surge in the diversity of hardware devices that NNs are deployed on. 
This presents a unique challenge as each device has its own attributes and the same NN may exhibit vastly different latency and energy characteristics across devices. 
Therefore, it becomes pivotal to co-optimize both accuracy and latency to meet stringent demands of real-world deployment~\cite{mnasnet, proxylessnas, fbnet, xu2020latency, cai2020once, wang2020hat}.
The simplest way to include latency optimization is to profile the latency of a NN on its target device.
However, this becomes costly and impractical when there are multiple small changes performed on the NN architecture either manually, or automatically through neural architecture search (NAS)~\cite{hwnassurvey}.
Not to mention that hardware devices are often not even available, or their software stacks do not yet support all NN architecture variants~\cite{bosch}, making it \textit{impossible} to perform hardware-aware NN optimizations.
For these reasons, much research has investigated statistical latency predictors that can accurately model hardware device latency with as few on-device measurements as possible.

\begin{figure*}[t!]
    \centering
    \caption{Predictors are central to hardware-aware neural architecture search, as they enable quick evaluation of candidate architectures. Latency predictors require several samples for training, but can be made more sample-efficient by first pre-training on a set of training devices. With the appropriate end-to-end latency predictor training pipeline, we can have extremely sample-efficient HW-Aware NAS!}
    \label{fig:teasermain}
    \includegraphics[width=0.9\linewidth]{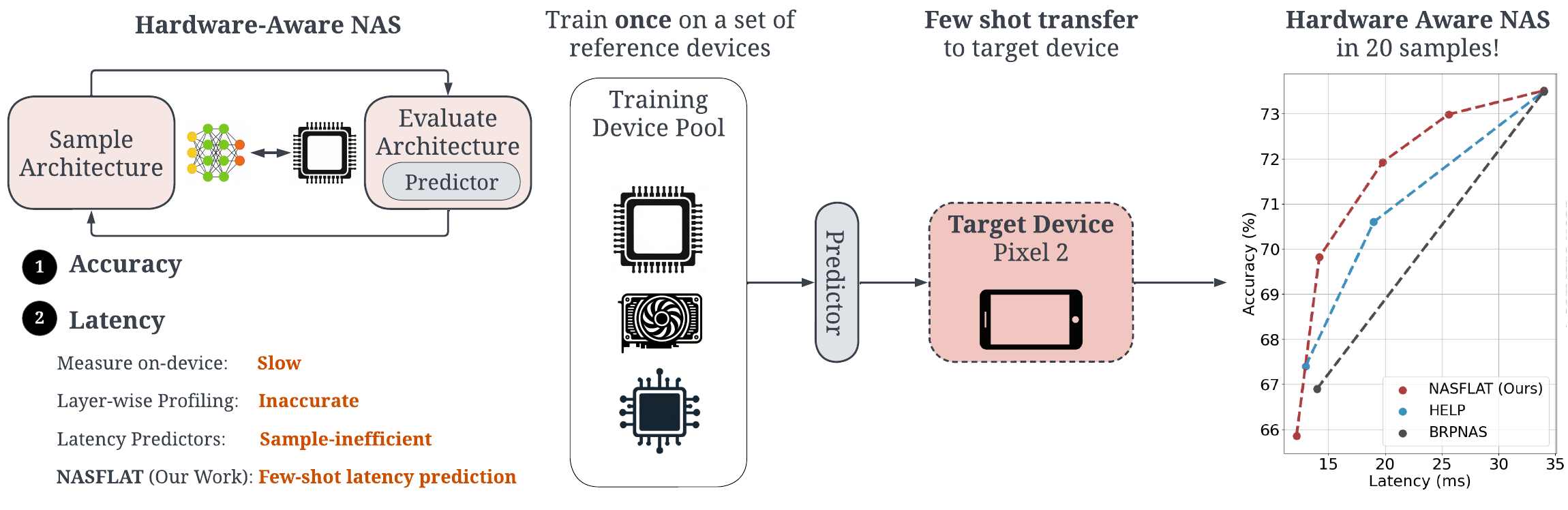}
\end{figure*}
\begin{figure*}[t!]
    \centering
    \caption{A few-shot latency predictor training pipeline. (1) We pre-train a predictor on a set of training devices. (2) The trained predictor can be adapted to any target device with just a few samples, using transfer learning. In this pipeline, the methodology used to sample neural networks as well as the predictor architecture play a key role.}
    \label{fig:overallflow}
    \includegraphics[width=\linewidth]{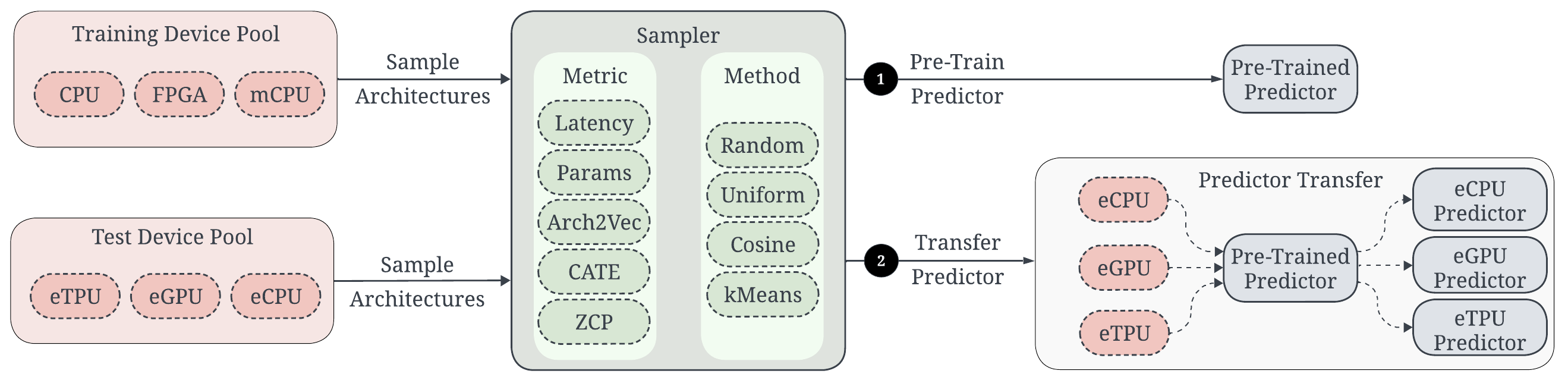}
\end{figure*}

Early work has used FLOPs as a proxy for latency~\cite{yu2020bignas}, while others created layerwise latency models~\cite{proxylessnas}.
However, both of these approaches generally performed poorly and could not adequately represent end-to-end device latency.
More recent work has used using Graph Convolutional Networks (GCNs)~\cite{brpnas} with much higher success in predicting the latency of NN architectures. 
However, a large number of NN latency samples needed to be gathered from each device for accurate modeling.
To mitigate this, HELP~\cite{help} and MultiPredict~\cite{multipredict} leverage transfer learning to train latency predictors with only a few NN latency samples. 
In this paradigm, a predictor is first trained to predict latency on a large set of \textit{training} devices (the training stage), and then through few-shot meta-learning, fine-tuned to predict on a set of \textit{test} devices (the transfer stage). 
This latter transfer stage can be performed efficiently using only a few sample measurements thus enabling the creation of latency predictors for new hardware devices in an inexpensive way.
Figure~\ref{fig:overallflow} illustrates the training and transfer of few-shot latency predictors.

This new class of few-shot latency predictors has become very practical and attractive for use within NAS and other NN latency optimization flows.
However, we have identified key shortcomings of existing works, as well as open research questions relating to the design of such predictors.
First, the choice of the few NN latency samples for transferring a predictor are critical.
Prior work has largely chosen this handful of NN architectures randomly but this often results in very high variance in predictive ability~\cite{help,multipredict}.
Simply put, the predictor performance is directly linked to the choice of those few samples.
Second, multi-hardware latency predictors require an additional input that represents the hardware device for both the pretraining and transfer phases.
One-hot encoding or a vector of latency measurements were used in the past for this purpose.
Third, NN operations were represented in the same way across different devices even though different operations exhibit different properties inherent to each device's hardware architecture and software compilation stack.
Finally, the predictor architecture itself was largely reused from prior work~\cite{brpnas} without explicit modifications for multi-device hardware predictors.
To address these main points, our work makes the following contributions:

\vspace{-.2cm}
\begin{enumerate}
\itemsep0em
    \item We investigate and empirically test different NN sampling methods for few-shot latency predictors, demonstrating a 5\% improvement compared to random sampling~\cite{help}, while requiring no additional samples, unlike uniform latency sampling~\cite{maple_edge}.
    \item We introduce hardware-specific NN operation embeddings to modulate NN encodings based on each hardware device, demonstrating a 7.8\% improvement. We additionally investigate the impact of supplementing with unsupervised (Arch2Vec) \cite{arch2vec}, computationally aware (CATE) \cite{cate}, and metric based (ZCP) \cite{abdelfattah2021zero} encodings resulting in 6.2\% improvement in prediction accuracy.
    \item Drawing from our evaluations on 12 experimental settings, we present \textbf{\GN{}}, \textbf{N}eural \textbf{A}rchitecture \textbf{S}ampler And \textbf{F}ew-Shot \textbf{Lat}ency Predictor, a multi-device latency predictor architecture which combines our graph neural network with an effective sampler, supplementary encodings and transfer learning to deliver an average latency predictor performance improvement of 22.5\%.
\end{enumerate}
\vspace{-.2cm}

Our detailed investigation offers insights into effective few-shot latency predictor design, and results in improvements up to 87.6\% on the most challenging prediction tasks (N2, FA, F2, F3 in Table \ref{table:all_final_results}), compared to prior work. Existing latency prediction tehniques often incur higher NAS costs owing to their sample inefficiency or complicated second-order transfer strategies. When evaluating end-to-end NAS outcomes, our approach demonstrates a $5.8\times$ speed-up in wall-clock time dedicated to latency predictor fine-tuning and prediction, compared to the best existing methods.



\comment{
\begin{itemize}
    \item \textit{Can encodings be used to improve predictors?} We investigate unsupervised (Arch2Vec) \cite{arch2vec}, computationally aware (CATE) \cite{cate}, and metric based (ZCP) \cite{multipredict,nasbenchsuitezero} methodologies for encoding neural networks, and its effectiveness as methods for sampling neural architecture during pre-training and transfer phases. We also study the effectiveness of such encodings as supplementary information to our latency predictor.
    \item \textit{Can we learn better hardware embeddings?} We introduce operation-specific HW embeddings to encode positional information about operations and capture interaction between layers.
    \item \textit{How can we design a good latency predictor?} We conduct an extensive ablation for latency predictor design, drawing from observations on NN encoding and HW embedding methodologies to design a state-of-the-art latency predictor that outperforms prior methods on 11 out of 12 HW latency prediction tasks. 
\end{itemize}
}

\section{Related Work}
\subsection{Hardware Latency Predictors}

A good latency predictor can enable NAS to co-optimize latency and accuracy~\cite{mnasnet, proxylessnas, fbnet, xu2020latency, cai2020once, wang2020hat}. 
Latency prediction methods have evolved from early, proxy based methods such as FLOPs \cite{yu2020bignas} to learning-based methods. 
This is largely because such proxies often do not correlate strongly with latency at deployment. 
To get better estimates for latency, some works used layer-wise latency prediction methods by measuring the latency for each operation and summing up the operations that the neural network has via a look-up table~\cite{proxylessnas}. This method does not account for operation pipelining or other compiler optimizations that may take place when multiple layers are executed consecutively.

BRP-NAS \cite{brpnas} takes into account such complexities and learns an end-to-end latency predictor which is trained on the target device. However, the latency measurements of a large number of NN architectures are required to perform well. This sample in-efficiency is largely because the latency predictor is trained from scratch. HELP \cite{help} employs a pool of reference devices to train its predictor and utilizes meta-learning techniques to adapt this predictor to a new device. The transfer of a predictor from some \textit{source} (training) devices to a \textit{target} (test) device significantly improves the sample efficiency of predictors. 
MultiPredict \cite{multipredict} facilitates predictor transfer across search spaces through unified encodings based on zero-cost proxies or hardware latency measurements.
Furthermore, MultiPredict investigates learnable hardware embeddings to represent different hardware devices within predictors. In our work, we extend the idea of a learnable hardware embedding to make it operation specific. Having a hardware embedding that explicitly interacts with the operation embeddings of a neural network architecture can capture intricacies in compiler level optimizations when executing hardware.

When transferring a predictor from a source device to a target device, the choice of samples used for few-shot learning plays a key role in the final performance. MAPLE-Edge\cite{maple_edge} investigates the impact of using the training device set latencies as reference for architecture to sample from the target device. For very large spaces, latencies of a sufficient diversity of neural network architectures may not be available even on the training devices. To address this, we look at methods of sampling a diverse set of neural networks from a target device which does not depend on latency or accuracy.

\subsection{Encodings for NN representation}
Early research in building predictors for accuracy and latency focused on using the adjacency and operation matrices to represent the directed acyclic graph (DAG) for the neural network architecture into a flattened vector to encode architectures \cite{encodingstudy}. HELP \cite{help} and MultiPredict \cite{multipredict} use the flattened one-hot operation matrix for the FBNet space with a multi layer perceptron for the latency and accuracy predictor. BRPNAS \cite{brpnas} employed a GCN with the adjacency-operation matrices as an input to build a latency and accuracy predictor for NASBench-201. More recently, works such as MultiPredict investigated the effect of capturing broad architectural properties by generating a vector of zero-cost proxies and hardware latencies to represent NNs. Additionally, there has been significant work in the field of encoding neural networks, notably the unsupervised learned encoding introduced in Arch2Vec \cite{arch2vec} which uses a graph auto-encoder to learn compressed latent representation for an NN architecture. Similarly, CATE \cite{cate} leveraged concepts from masked language modeling to learn encodings for computationally similar architectures with a transformer. 

In our work, we leverage these NN encodings to sample diverse architectures in the neural architecture search space. We also leverage these encodings to provide additional architectural information to our latency predictor.

\begin{figure*}[t!]
    \centering
    \caption{Our model architecture maintains separate operation and hardware embeddings which are concatenated and passed to a GNN to refine and contextualize the embeddings with respect to the overall neural architecture. This serves as the architecture embedding and is passed to another GNN along with the adjacency and node information. Optionally, supplementary encodings can be concatenated with the output of the GNN and fed to a prediction head to estimate the latency of the architecture.}
    \label{fig:model_arch}
    \includegraphics[width=.9\linewidth]{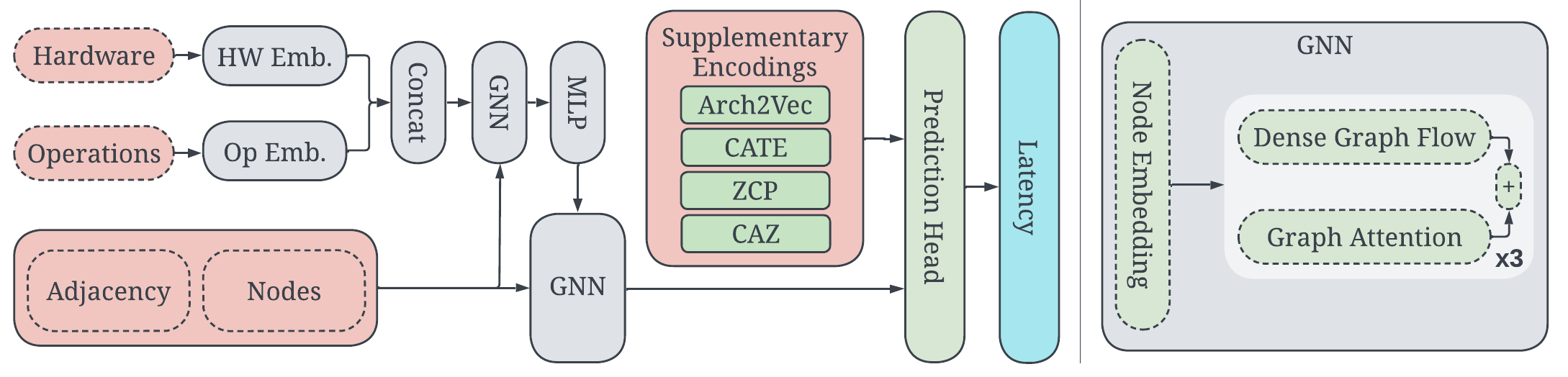}
\end{figure*}

\section{Latency Predictor Design}
\label{sec:latpreddesign}


\subsection{Predictor Architecture}
The design of the predictor itself plays a key role in improving sample efficiency of accuracy predictors in NAS. 
BRP-NAS~\cite{brpnas} used a GCN predictor to capture information about both the NN operations and connectivity.
The same predictor has subsequently been used by HELP~\cite{help}.
TA-GATES \cite{tagates} further enhanced this predictor architecture with residual connections in the GCN module and a \textit{training-analogous} operation update methodology, by maintaining a backward graph neural network module. 
This allows the iterative refinement of operation embeddings to provide more information about the architecture. 
We investigate the impact of these and other key components of predictor design in the Appendix, and design a latency predictor that accounts for both operation and hardware embeddings.
We subsequently use this predictor for all experiments in this paper.

Figure \ref{fig:model_arch} shows how we maintain separate embedding tables for the hardware device and operations. 
From our ablations in the Appendix, we find that a single GNN module to refine the operation embeddings is sufficient. 
To capture the complex interactions such as layer pipelining, operator fusion, we further incorporate the hardware embedding into each of the operation embeddings via concatenation. 
This joint embedding is then passed to a small operation GNN along with the adjacency and node embedding. 
Then, the embedding of the output node of the GNN is provided to an MLP which provides the hardware-operation joint embedding for the second (main) GNN used for modeling an input NN. 
Finally, supplementary encoding (discussed more below) can be concatenated with the GNN output and provided to a prediction head that can then output the latency metric.

\subsection{GNN Module Design}
Existing Graph Convolutional Networks (GCNs) experience an over-smoothing challenge, leading to a loss of discriminative information in the node embedding as aggregation layers increase. 
Addressing this, GATES~\cite{gates} introduced the Dense Graph Flow (DGF) module, which employs residual connections to maintain discriminative features across nodes.  Furthermore, the Graph Attention (GAT) methodology, distinct from DGF, incorporates an attention mechanism during node aggregation. GAT evaluates node interactions through an attention layer. An operation attention mechanism, along with LayerNorm, refines information aggregation and ensures stable training, respectively. Further details of their implementation are in the Appendix. In our latency predictor, we use an ensemble of DGF and GAT modules. 

\subsection{Supplementary Encodings}

Supplementary encodings are different ways to represent the input NN, and therefore may help contextualize the relations between a NN with respect to the entire search space.
This can be useful for few-shot transfer of latency predictors.
Learned encodings like Arch2Vec~\cite{arch2vec}, ZCP~\cite{multipredict} and CATE~\cite{cate} provide distinct representations that allows them to effectively distinguish between various neural network (NN) architectures. 
For example, CATE \citep{cate} captures computational characteristics of NNs through its latent representations formed by computational clustering. Simultaneously, ZCP offers insights at the architectural level, acting as proxies that might correlate with accuracy.

To enhance the robustness and accuracy of our latency predictor, we integrate these encodings into its structure. Specifically, Arch2Vec, CATE, and ZCP encodings are introduced as supplementary inputs to the predictor head of our latency predictor. As illustrated in Figure~\ref{fig:model_arch}, these encodings are fed into the MLP prediction head subsequent to the node aggregation phase. This augmentation not only incorporates the rich structural and computational characteristics of NN architectures but also helps the predictor to make better-informed latency estimates. In our experiments, we observed that incorporating architectural-level information from encodings can boost the sample-efficiency of our latency predictors. We also introduce the CAZ encoding, a combined representation formed by concatenating CATE, Arch2Vec, and ZCP, aiming to leverage the combined strengths of all three representations.

\subsection{Transfer Of Pre-Trained Predictor}
To pretrain the predictor, we form a large dataset from a number of source devices. This conventional training step is the same as prior work~\cite{help,multipredict}.
Once this pre-training phase is concluded, the subsequent step is the adaptation to a target hardware device. In alignment with the methodology described in MultiPredict~\cite{multipredict}, the predictor undergoes a fine-tuning process using the samples from the target device. 
The learning rate is re-initialized and fine-tuning of the predictor is conducted on the target device. We find that this is sufficient to calibrate the predictor to offer accurate latency estimates on the unseen device.


\section{Neural Network Samplers}

One of the key aspects of latency predictor training, especially in the low-sample count regime is choosing diverse neural networks. 
If all the samples profiled on the target device have similar computational characteristics, the predictor may not gather enough information to generalize. 
As depicted in Figure \ref{fig:overallflow}, one of the key aspects of predictor training and transfer is the sampler, which needs to select a diverse set of neural architectures to benchmark for few-shot learning. 
MAPLE-Edge \cite{maple_edge} uses latencies on a set of reference devices to identify architectures with distinct computational properties. 
While this is a very effective strategy to identify computationally distinct neural networks, this requires a very large number of on-device latency measurements---a key parameter that we would like to minimize. 

In this section, we investigate methods of encoding NN architectures. We look at the impact of these encodings in helping us sample more diverse architectures. One key benefit of using these encodings to sample diverse architectures is that we no longer depend on reference device latency measurements.

\subsection{Neural Network Encodings}

A foundational aspect of hardware-aware NAS is to optimize an objective function $\ell:A\rightarrow\mathbb{R}$, where $\ell$ can quantify several performance metrics such as accuracy, latency, energy \cite{brpnas}. $A$ represents the NN search space, and architectures $a \in A$ can be represented as adjacency - operation matrices \cite{encodingstudy}. 
There are several methods that introduce alternative methods to represent $a$ \cite{arch2vec, cate, multipredict}. In this section, we look at some of these methods of encoding neural networks.

\textbf{Learned} encodings aim to represent the structural properties of a neural architecture in a latent vector without utilizing accuracy. Arch2Vec \cite{arch2vec} uses a variational graph isomorphism autoencoder to learn to regenerate the adjacency-operation matrix. CATE \cite{cate} introduces a transformer that uses computationally similar architecture pairs (clustered by similar FLOPs or parameter count) to learn encodings. Naturally, the CATE encoding clusters architectures that have similar computational properties. 

\textbf{Zero-Cost Proxies (ZCP)} encode neural networks as a vector of metrics. Each of these metrics attempt to encode properties of a neural network that may correlate with accuracy. Distinct connectivity patterns and operation choices (via varying adjacency-operation matrices) would initialize NNs that exhibit varied accuracy and latency characteristics. Thus, ZCP can implicitly capture architectural properties of neural networks, but does not contain explicit structural information.

\subsection{Encoding-based Samplers}

Encodings like ZCP, Arch2Vec, and CATE condense a broad spectrum of architectural information into a latent space. While Arch2Vec compresses the adjacency-operation matrix, capturing its intrinsic structure, CATE identifies and groups computationally similar architectures. In contrast, ZCPs capture global properties of the neural network that may correlate with accuracy or may encode operator level information about the architecture. Such encodings, collectively contain a rich representation of the entire neural architecture design space which can be used to decide which architectures to obtain the latency for on the target device.

For the ZCP encoding, we use 13 zero cost proxies, and generate 32-dimensional vectors for the Arch2Vec and CATE encodings. We further introduce an encoding `CAZ', which combines the CATE, Arch2Vec and ZCP. Given the richness and diversity of the encoded representation of neural architectures, a systematic approach to selection becomes essential. We thus investigate two methodologies of using the encoding to identify architectures to sample for few-shot transfer to a target device.

\textbf{Cosine Similarity and KMeans Clustering}: Through our framework, we leverage cosine similarity \cite{cosinesimilarity}—an intuitive metric of vector similarity—to discern architectures that may have distinct properties. Focusing on structures with reduced average cosine similarities ensures a wider design space coverage, potentially identifying `outlier' architectures. Concurrently, utilizing the KMeans clustering algorithm \cite{kmeanspaper}, we categorize encoded vectors into distinct groups, and opt for the one closest to the centroid of each cluster. The rationale is that these architectures are most representative of their respective clusters and hence provide a good spread across the design space.

\section{Hardware Embeddings}
\label{sec:hwembedding}


HELP and MultiPredict \cite{help, multipredict} investigate different methods of representing hardware, as an assigned device index, a vector of architectural latency measurements, or as a learnable hardware embedding table, which is relayed to the predictor for identifying devices. However, such an approach potentially oversimplifies the intricate dynamics of neural network deployment on hardware, thereby introducing the need for an interaction between the operation and hardware embedding. In this section, we discuss a methodology to initialize and utilize hardware embeddings to better model the dynamics of the target hardware.

\begin{algorithm}[t!]
   \caption{Methodology to partition device sets.}
   \label{alg:bipartitePrune}
\begin{algorithmic}
   \STATE {\bfseries Input:} Graph $G$ (negative correlations), integers $m$, $n$
   \STATE {\bfseries Output:} Modified graph $B$

    \STATE $b\_m, b\_n = \texttt{kernighan-lin(G)}$
    \STATE \algorithmiccomment{Initialize bipartite graph with correlations}
    \STATE \( B = \texttt{initBipartite}(b\_m, b\_n, \texttt{correlations}) \)
    \WHILE{\texttt{len(B[0])} $\neq$ m \OR \texttt{len(B[1])} $\neq$ n}
    \STATE \algorithmiccomment{Identify disjoint device sets $U$ and $V$}
    \STATE \( \texttt{l, r} = \texttt{B[0], B[1]} \)

    \STATE \algorithmiccomment{Remove node with highest correlation.}
    \IF{\texttt{len(B[0])} $>$ m}
    \STATE \texttt{removeMaxWeightNode(B, l)}
    \ENDIF
    \IF{\texttt{len(B[1])} $>$ n}
    \STATE \texttt{removeMaxWeightNode(B, r)}
    \ENDIF
    \ENDWHILE
\end{algorithmic}
\end{algorithm}

\subsection{Operation Specific Hardware Embedding}
From a hardware perspective, the location of an operation in relation to its preceding and succeeding layers can considerably influence overall latency. This can be attributed to optimizations such as layer pipelining, where operations are organized in a staggered manner to maximize hardware utilization. Additionally, optimizations such as layer fusion which combine adjacent layers to streamline computations further underscore the importance of operation placement. 

The latency predictor accepts the adjacency matrix, node, and operation indices as its inputs. 
The operation index is utilized to retrieve a learnable operation embedding from an embedding table, which encodes the properties and behaviors of the respective operation, however, when modeling latencies for various hardware devices, this singular operation embedding might not fully capture the nuances of each hardware. As depicted in Figure \ref{fig:model_arch}, we concurrently incorporate hardware-specific embeddings into our predictor, such that we are able to model the interaction between an operation and the hardware depending on its nature and position. Thus, the operation-specific hardware embedding introduces a more granular approach wherein each operation within the neural network is concatenated with a specific hardware embedding from an embedding table. Such a strategy not only encapsulates the intrinsic characteristics of the operation but also embeds information regarding how that specific operation would behave on the given hardware.

\begin{table}[t!]
  \centering
\caption{Device sets for NASBench-201 and FBNet. S, T indicate source and target device pools respectively. Each \textit{device pool} may contain more than one device, full details in the Appendix. Unless otherwise specified, we report the \textit{average} correlation and standard deviation across trials \textit{and} target (T) devices.}\vspace{2mm}
\label{table:datasets}
  \resizebox{1.\linewidth}{!}{%
\centering
\begin{tabular}{ccccccccccccc}
\toprule
 \multirow{2}{*}{Devices} & \multicolumn{6}{c}{NB201} & \multicolumn{6}{c}{FBNet} \\  \cmidrule(lr){2-7} \cmidrule(lr){8-13} \addlinespace[0.5ex]
 & ND & N1 & N2 & N3 & N4 & NA & FD & F1 & F2 & F3 & F4 & FA \\ \cmidrule(lr){1-1} \cmidrule(lr){2-7} \cmidrule(lr){8-13} \addlinespace[0.5ex]
DSP & - & - & T & - & - & - & - & - & - & - & - & - \\ 
CPU & ST & - & - & S & S & ST & ST & S & S & - & S & T \\ 
GPU & ST & T & S & T & T & S & ST & S & T & T & ST & S \\ 
FPGA & T & - & - & - & - & - & - & T & S & S & - & - \\ 
ASIC & T & S & - & S & S & T & - & T & - & - & S & - \\ \cmidrule(lr){1-1} \cmidrule(lr){2-7} \cmidrule(lr){8-13} \addlinespace[0.5ex]
eTPU & - & S & T & - & S & T & - & - & - & - & - & - \\ 
eGPU & - & - & T & S & S & - & - & - & - & - & - & - \\ 
eCPU & T & - & T & - & - & - & S & T & - & - & S & - \\ \cmidrule(lr){1-1} \cmidrule(lr){2-7} \cmidrule(lr){8-13} \addlinespace[0.5ex]
mGPU & - & S & - & S & S & - & - & - & - & - & - & - \\ 
mCPU & S & S & - & - & S & S & - & S & S & S & ST & T \\ 
mDSP & - & - & - & S & S & - & - & - & - & - & - & - \\ \bottomrule
\end{tabular}%
}
\end{table}

\begin{table*}[t!]
    \centering
  \caption{The impact of operation-wise hardware embedding (OPHW) and hardware embedding initialization (INIT) on latency predictor performance. Both optimizations consistently demonstrate an improvement.}\vspace{2mm}
  \label{table:opwise_emb}
    \resizebox{1\linewidth}{!}{%
  \centering
  
\begin{tabular}{ccccccccccccc}
\toprule
OPHW    &       ND  &       N1  &       N2  &       N3  &       N4   &      NA  &    FD  &       F1  &       F2  &       F3  &       F4  &       FA \\
\cmidrule(lr){1-1} \cmidrule(lr){2-7}\cmidrule(lr){8-13} \addlinespace[0.5ex]
\ding{55}  &       $0.804_{0.026}$ & $0.701_{0.056}$ & $0.763_{0.044}$ & $0.680_{0.042}$ & \bR$0.757_{0.061}$ & $0.660_{0.030}$&  $0.839_{0.010}$ & $0.753_{0.031}$ & $0.745_{0.081}$ & $0.685_{0.085}$ & $0.813_{0.030}$ & $0.566_{0.081}$ \\
\ding{51} &       \bR$0.806_{0.038}$ & \bR$0.719_{0.050}$ & \bR$0.795_{0.034}$ & \bR$0.704_{0.058}$ & $0.753_{0.052}$ & \bR$0.664_{0.059}$ &        \bR$0.845_{0.009}$ & \bR$0.757_{0.048}$ & \bR$0.769_{0.076}$ & \bR$0.694_{0.076}$ & \bR$0.832_{0.026}$ & \bR$0.628_{0.103}$ \\
\addlinespace[0.5ex]\toprule\addlinespace[0.5ex]
INIT    &       ND  &       N1  &       N2  &       N3  &       N4   &      NA  &       FD  &       F1  &       F2  &       F3  &       F4  &       FA \\
\cmidrule(lr){1-1} \cmidrule(lr){2-7}\cmidrule(lr){8-13}\addlinespace[0.5ex]
\ding{55} & $0.794_{0.038}$ & $0.701_{0.046}$ & $0.791_{0.036}$ & \bR$0.718_{0.055}$ & $0.750_{0.050}$ & $0.658_{0.080}$ &  $0.707_{0.158}$ & $0.609_{0.119}$ & $0.754_{0.064}$ & $0.655_{0.084}$ & $0.659_{0.158}$ & $0.559_{0.118}$ \\
\ding{51} &       \bR$0.806_{0.038}$ & \bR$0.719_{0.050}$ & \bR$0.795_{0.034}$ & $0.704_{0.058}$ & \bR$0.753_{0.052}$ & \bR$0.664_{0.059}$ &  \bR$0.845_{0.009}$ & \bR$0.757_{0.048}$ & \bR$0.769_{0.076}$ & \bR$0.694_{0.076}$ & \bR$0.832_{0.026}$ & \bR$0.628_{0.103}$ \\
\bottomrule
\end{tabular}%
}
\end{table*}

\begin{table*}[t!]
  \centering
\caption{On over 10 out of 12 device sets, there is a benefit in using learned or zero-cost encodings for training-transfer sample selection.}\vspace{2mm}
\label{table:sampler_study}
  \resizebox{0.7\linewidth}{!}{%
\begin{tabular}{lllllll}
\toprule
Sampler &                ND &                N1 &                N2 &                N3 &                N4 &                NA \\
\cmidrule(lr){1-1} \cmidrule(lr){2-7} \addlinespace[0.5ex]
Latency (Oracle)          &      $0.929_{0.027}$  &      $0.960_{0.011}$ &      $0.793_{0.078}$ &      $0.951_{0.021}$ &      $0.919_{0.055}$   &      $0.851_{0.039}$ \\
\cmidrule(lr){1-1} \cmidrule(lr){2-7}  \addlinespace[0.5ex]
Random              &      $0.911_{0.038}$  &      $0.946_{0.026}$ &      $0.757_{0.052}$ &      $0.934_{0.032}$ &  \bB $0.940_{0.026}$   &      $0.790_{0.070}$ \\
Params              &      $0.898_{0.068}$  &      $0.934_{0.038}$ &  \bR $0.767_{0.068}$ &      $0.918_{0.033}$ &      $0.940_{0.032}$   &  \bB $0.801_{0.095}$ \\
Arch2Vec            &  \bB $0.912_{0.045}$  &      $0.931_{0.046}$ &      $0.741_{0.073}$ &      $0.930_{0.036}$ &      $0.907_{0.069}$   &  \bR $0.849_{0.035}$ \\
CATE                &      $0.893_{0.045}$  &      $0.937_{0.036}$ &      $0.761_{0.090}$ &  \bB $0.935_{0.032}$ &  \bR $0.945_{0.038}$   &      $0.767_{0.136}$ \\
ZCP                 &      $0.883_{0.075}$  &  \bB $0.956_{0.039}$ &      $0.636_{0.170}$ &      $0.924_{0.051}$ &      $0.883_{0.071}$   &      $0.729_{0.212}$ \\
CAZ                 &  \bR $0.925_{0.046}$  &  \bR $0.957_{0.028}$ &  \bB $0.761_{0.107}$ &   \bR$0.935_{0.025}$ &      $0.866_{0.091}$   &      $0.680_{0.336}$ \\
\cmidrule(lr){1-1} \cmidrule(lr){2-7} \addlinespace[0.5ex]
Sampler &           FD &                F1 &                F2 &                F3 &                F4 &                FA \\
\cmidrule(lr){1-1} \cmidrule(lr){2-7}\addlinespace[0.5ex]
Latency (Oracle)          &      $0.755_{0.048}$ &      $0.707_{0.052}$ &      $0.832_{0.035}$ &      $0.849_{0.022}$ &      $0.699_{0.077}$  &      $0.624_{0.112}$\\
\cmidrule(lr){1-1} \cmidrule(lr){2-7}  \addlinespace[0.5ex]
Random            &      $0.665_{0.187}$ &      $0.642_{0.121}$ &   \bB$0.801_{0.063}$ &  \bR $0.809_{0.050}$ &      $0.658_{0.113}$  &  \bB $0.615_{0.115}$\\
Params            &      $0.735_{0.073}$ &      $0.689_{0.070}$ &      $0.794_{0.078}$ &      $0.791_{0.062}$ &      $0.604_{0.239}$  &      $0.551_{0.146}$\\
Arch2Vec          &  \bR $0.754_{0.071}$ &  \bR $0.699_{0.046}$ &      $0.790_{0.065}$ &      $0.782_{0.083}$ &  \bR $0.739_{0.054}$  &  \bR $0.631_{0.169}$\\
CATE              &      $0.663_{0.132}$ &  \bB $0.692_{0.079}$ &      $0.778_{0.078}$ &      $0.780_{0.076}$ &      $0.645_{0.118}$  &      $0.552_{0.147}$\\
ZCP               &  \bB $0.744_{0.060}$ &      $0.665_{0.111}$ &      $0.789_{0.069}$ &  \bB $0.801_{0.055}$ &  \bB $0.734_{0.051}$  &      $0.586_{0.161}$\\
CAZ               &      $0.696_{0.107}$ &      $0.635_{0.105}$ &  \bR $0.808_{0.040}$ &      $0.732_{0.097}$ &      $0.626_{0.127}$  &      $0.557_{0.141}$\\
\bottomrule
\end{tabular}
}
\end{table*}

\subsection{Hardware Embedding Initialization}
When adding a new target device, a good initialization for its hardware embedding is critical. 
For this, we gauge the computational correlation of the target device latency with each of the training devices. 
By identifying the training device with the highest correlation, we can use its learned hardware embedding as the starting point for the target device. 
This method harnesses latency similarities between devices, providing a good initialization for predictions on the target device. 
In addition, it avoids a \textit{cold start} when using the predictor for a new device, allowing the predictor to be functional with just a small number of latency samples on the new device.

\section{Experiments}

\begin{table*}[t!]
  \centering
\caption{Over three device sets on two NAS spaces, there is a benefit in using supplementary encodings for representing neural networks.}\vspace{2mm}
\label{table:adj_gin_enc_study}
  \resizebox{0.7\linewidth}{!}{%
\begin{tabular}{lllllll}
\toprule
Encoding &                ND &                N1 &                N2 &                N3 &                N4 &                NA  \\
\cmidrule(lr){1-1} \cmidrule(lr){2-7} \addlinespace[0.5ex]
AdjOp         &  $0.959_{0.018}$ &      $0.971_{0.010}$ &      $0.848_{0.063}$ &      $0.966_{0.006}$ &      $0.965_{0.012}$ &      $0.893_{0.029}$ \\
\cmidrule(lr){1-1} \cmidrule(lr){2-7}  \addlinespace[0.5ex]
(+ Arch2Vec) &  \bR $0.961_{0.010}$ &  \bR $0.972_{0.007}$ &      $0.816_{0.065}$ &  \bR $0.968_{0.005}$ &      $0.965_{0.009}$ &      $0.895_{0.035}$ \\
(+ CATE)     &      $0.954_{0.022}$ &      $0.962_{0.029}$ &      $0.833_{0.072}$ &   \bB$0.967_{0.007}$ &  \bR $0.967_{0.008}$ &  \bR $0.907_{0.024}$ \\
(+ ZCP)      &      $0.955_{0.024}$ &      $0.968_{0.014}$ &   \bB$0.855_{0.026}$ &      $0.963_{0.009}$ &      $0.962_{0.011}$ &      $0.896_{0.018}$ \\
(+ CAZ)      &   \bB$0.960_{0.014}$ &   \bB$0.972_{0.005}$ &  \bR $0.861_{0.036}$ &      $0.965_{0.005}$ &   \bB$0.966_{0.006}$ &  \bB $0.899_{0.012}$ \\
\cmidrule(lr){1-1} \cmidrule(lr){2-7}  \addlinespace[0.5ex]
Encoding &                FD &                F1 &                F2 &                F3 &                F4 &                FA \\
\cmidrule(lr){1-1} \cmidrule(lr){2-7} \addlinespace[0.5ex]
AdjOp          &      $0.783_{0.035}$ &      $0.744_{0.032}$ &      $0.855_{0.021}$ &      $0.850_{0.026}$ &      $0.750_{0.066}$ &      $0.694_{0.060}$ \\
\cmidrule(lr){1-1} \cmidrule(lr){2-7} \addlinespace[0.5ex]
(+ Arch2Vec) &      $0.881_{0.016}$ &      $0.788_{0.027}$ &  \bB $0.878_{0.020}$ &  \bB $0.890_{0.012}$ &  \bR $0.848_{0.020}$ &  \bB $0.723_{0.034}$ \\
(+ CATE)     &      $0.837_{0.031}$ &      $0.744_{0.037}$ &      $0.839_{0.022}$ &      $0.845_{0.037}$ &      $0.805_{0.023}$ &      $0.678_{0.055}$ \\
(+ ZCP)      &  \bR $0.960_{0.008}$ &  \bR $0.842_{0.020}$ &  \bR $0.895_{0.018}$ &  \bR $0.899_{0.019}$ &  \bB $0.843_{0.028}$ &  \bR $0.776_{0.038}$ \\
(+ CAZ)      & \bB  $0.899_{0.021}$ &  \bB $0.783_{0.033}$ &      $0.842_{0.036}$ &      $0.852_{0.029}$ &      $0.761_{0.056}$ &      $0.683_{0.077}$ \\
\bottomrule
\end{tabular}
}
\end{table*}

In this section, we systematically investigate key design considerations for predictor design. We begin by looking at the effectiveness of our proposed operation specific hardware embedding as well as hardware embedding initialization. We further investigate the impact that graph neural network module design has on predictor efficacy, looking at graph convolutional networks, graph attention networks and their ensemble. We then study the impact of neural network encoding strategies on selection of neural network architectures for transferring predictors to a target device. 
By supplementing our predictor with additional NN encodings, we provide additional information coveying the relative performance of NNs within a search space---our ablation shows improved prediction.
Finally, we combine our empirically-driven optimizations on sampling, encodings, hardware embeddings, and predictor architecture to design a state-of-the-art latency predictor.
Our evaluation encompasses both the Spearman Rank Correlation of predicted latency and ground truth on multiple device sets, in addition to end-to-end HW-Aware NAS. 


\subsection{Designing Evaluation Sets}


One of the key challenges in evaluating HW-Aware NAS is the lack of standardized evaluation sets and hardware latency data-sets. HW-NAS-Bench \cite{hwnasbench}, HELP \cite{help} and EAGLE \cite{brpnas} collectively open-source latencies for a wide range of hardware on the NASBench-201 and FBNet design  spaces, making HW-Aware NAS evaluation easier. However, MultiPredict identified an issue where device sets on which current latency predictors were evaluated on showed high training-test correlation. For NASBench-201 and FBNet, high training-test correlation device sets are denoted as `ND' and `FD', respectively. 
To this end, MultiPredict introduced device sets with low training-test correlation to evaluate latency predictors, indicated by `NA' and `FA' (\textit{A} signifying the adversarial nature of the device set) in Table \ref{table:datasets}. A key shortcoming in these existing works is that the devices that the predictor is evaluated on is cherry-picked, exhibiting a high correlation between the training devices and test devices, a property that is not guaranteed in practice. To circumvent this limitation, we employ an automated algorithmic strategy to design training and test devices to maintain low mutual correlation. Initially, we compute the Spearman correlation of latencies between all available devices, and construct a graph structure with negative correlations between devices which serve as edge weights. As depicted in Algorithm \ref{alg:bipartitePrune}, we then leverage the Kernighan-Lin \cite{kernighan_len_bipartite} bisection method to bisect the graph, aiming to group devices with minimal intra-group correlation. We iteratively trim the bipartite graph to maintain a specified number of devices in each set. By algorithmically partitioning device sets, we ensure an objective and unbiased selection process. With this strategy, we introduce four device sets for NASBench-201 and FBNet, identified by (N1, N2, N3, N4) and (F1, F2, F3, F4) in Table \ref{table:datasets}.

\subsection{Experimental Setup}

In each of our experiments, we pretrain our predictor on many samples from multiple source devices, then we fine-tune the predictor on only a few samples from the target devices as defined by our evaluation sets.
We report the Spearman Rank Correlation Coefficient of predicted latency relative to ground-truth latency values as a measure of predictive performance. In Figure \ref{table:gcn_gat_design}, we investigate the GAT and DGF GNN module. In all our experiments, we decide to use the DGF-GAT ensemble as the GNN module. The appendix details the precise experimental settings for each of our results.

\begin{table}[t!]
  \centering
\caption{On NB201, GAT outperforms DGF. The difference is less evident on FBNet, we thus use an ensemble of DGF and GAT.}\vspace{2mm}
\label{table:gcn_gat_design}
  \resizebox{\linewidth}{!}{%
\begin{tabular}{ccccc}
\toprule
GNN Module & ND & N1 & N2 & N3  \\
\cmidrule(lr){1-1} \cmidrule(lr){2-5} \addlinespace[0.5ex]
DGF       & $0.814_{0.026}$ & $0.741_{0.032}$ & \bR$0.802_{0.021}$ & $0.710_{0.042}$  \\ 
GAT       & \bR$0.965_{0.005}$ & \bR$0.848_{0.038}$ & $0.612_{0.032}$ & $0.762_{0.039}$ \\ 
Ensemble  & $0.965_{0.011}$ & $0.829_{0.046}$ & $0.634_{0.029}$ & \bR$0.789_{0.054}$  \\ 
\cmidrule(lr){1-1} \cmidrule(lr){2-5} \addlinespace[0.5ex]
GNN Module & FD & F1 & F2 & F3 \\
\cmidrule(lr){1-1} \cmidrule(lr){2-5} \addlinespace[0.5ex]
DGF       &\bR$0.844_{0.004}$ & $0.740_{0.055}$ & $0.752_{0.076}$ & \bR$0.621_{0.120}$ \\ 
GAT       &  $0.823_{0.019}$ & \bR$0.749_{0.042}$ & $0.700_{0.114}$ & $0.589_{0.083}$ \\ 
Ensemble  & $0.835_{0.007}$ & $0.733_{0.059}$ & \bR$0.766_{0.013}$ & $0.609_{0.106}$ \\ 
\bottomrule
\end{tabular}%
}
\end{table}

\subsection{Hardware-aware Operation Embeddings}
Instead of representing operations with a fixed embedding, we modulate embeddings based on each hardware device as described in Section~\ref{sec:hwembedding}.
Table \ref{table:opwise_emb} evaluates the impact of this optimization on the predictive ability using our latency predictor.
We find that on 11 out of 12 device pools, there is a positive impact of introducing the operation-wise hardware embedding. 
Additionally, we evaluate the impact of initializing the hardware embedding of the target device with the embedding of the most closely-correlated source device. 
Our results show consistent improvement from this initialization scheme as shown in Table~\ref{table:opwise_emb}.
These two optimizations have empirically demonstrated their efficacy in utilizing hardware-specific operation embeddings, and mitigating new device cold start in our predictor.

\begin{table*}[t!]
  \centering
\caption{We study the impact of different design choices on the performance of predictors. Each row adds a feature and also inherits the design choices above it.}\vspace{2mm}
\label{table:design_components}
  \resizebox{\linewidth}{!}{%
\begin{tabular}{lcccccccc}
\toprule
                        & F1  & F2 & F3   &   F4 & N1 & N2 & N3 & N4 \\
  \cmidrule(lr){1-1} \cmidrule(lr){2-5} \cmidrule(lr){6-9} \addlinespace[0.5ex]
Baseline Predictor     & $0.603_{0.104}$ & $0.800_{0.050}$ &    $0.792_{0.058}$  &      $0.502_{0.142}$      & $0.938_{0.034}$ & $0.781_{0.084}$  & $0.907_{0.019}$ & $0.922_{0.021}$ \\
(+ HWInit)             & $0.573_{0.114}$ & $0.770_{0.072}$ &    \bB$0.822_{0.053}$  &      $0.566_{0.127}$       & $0.938_{0.031}$ & $0.776_{0.039}$  & $0.887_{0.087}$ & $0.898_{0.046}$ \\
(+ Op$_{HW}$)          & $0.610_{0.076}$ & $0.801_{0.045}$ &    \bR$0.824_{0.025}$  &      $0.549_{0.078}$    & \bB$0.949_{0.017}$ & \bR$0.821_{0.045}$  & \bR$0.938_{0.021}$ \bB& $0.955_{0.015}$ \\
(+ Sampler)            & \bB$0.639_{0.069}$ & \bB$0.813_{0.042}$ &    $0.810_{0.053}$  &      \bB$0.616_{0.116}$      & \bR$0.962_{0.012}$ & $0.803_{0.067}$  & $0.920_{0.034}$ & \bR$0.961_{0.007}$ \\
(+ Supp. Encoding)     & \bR$0.727_{0.048}$ & \bR$0.844_{0.031}$ &    $0.816_{0.057}$  &      \bR$0.796_{0.045}$     & $0.936_{0.033}$ & \bB$0.812_{0.059}$  & \bB$0.930_{0.028}$ & $0.940_{0.017}$ \\
                        
\bottomrule

\end{tabular}%
}
\end{table*}

\subsection{Encoding-based Samplers}

Here, we evaluate different sampling methods, specifically those based on uniformly sampling NNs based on different encodings.
Our findings in Table~\ref{table:sampler_study} present a somewhat varied pattern. While encodings-based samplers were advantageous in 10 out of 12 device pools, determining the optimal sampler became a complex task dependent on the each device pool. 
Notably, CATE's performance was subpar, especially on FBNet. A potential reason for this could be the disparity between the FBNet search space, which possesses $9^{22}$ unique architectures, and the latency dataset available in HWNASBench, restricted to only $5000$ architectures. 
Consequently, when training the CATE encoding on this limited set, computationally similar architectures do not carry much meaning with respect to the entire FBNet search space. 
Arch2Vec is trained similarly, but since computationally similar architectures are not required, it is able to learn a representation on a smaller space.

Delving deeper into the selection strategy used for clustering the encodings, we observed a pronounced inclination towards the cosine similarity approach. As detailed in the appendix, Cosine consistently demonstrated superior performance over KMeans. Additionally, KMeans was occasionally unable to segment the space adequately to yield architectures (as evidenced by NaN entries).


\subsection{Supplementary NN Encodings}

\begin{table*}[t!]
  \centering
\caption{We train our predictor with our proposed sampler, GAT+GCN ensemble architecture, operation-wise hardware embedding, hardware embedding initialization, and report our end-to-end predictor transfer result (Spearman Rank Correaltion). On 11 out of 12 tasks, our predictor works best for NASBench-201 and FBNet respectively. Standard deviations are reported for results we produce in our paper.}\vspace{2mm}
\label{table:all_final_results}
  \resizebox{0.8\linewidth}{!}{%
\begin{tabular}{cccccccccccccccc}
\toprule
  & Source & Target             & \multirow{2}{*}{ND} & \multirow{2}{*}{NA} & \multirow{2}{*}{N1} & \multirow{2}{*}{N2} & \multirow{2}{*}{N3} & \multirow{2}{*}{N4} & \multirow{2}{*}{GM} \\
  & \multicolumn{2}{c}{Samples} &                     &                     &                     &                     &                     &                     &   \\
  \cmidrule(lr){1-1} \cmidrule(lr){2-3} \cmidrule(lr){4-10} \addlinespace[0.5ex]
HELP                & 900 &  20 & $0.948_{0.006}$ & $0.410_{0.037}$ & $0.604_{0.044}$ & $0.509_{0.007}$ & $0.729_{0.027}$ & $0.746_{0.042}$ & $0.634$\\
MultiPredict        & 900 &  20 & $0.930_{0.012}$ & $0.820_{0.019}$ & $0.907_{0.003}$ & $0.757_{0.045}$ & $0.947_{0.012}$ & $0.952_{0.011}$  &$0.882$\\
            \textbf{\GN{}}   & \bR$25$  &  20 & \bR$0.959_{0.007}$ & \bR$0.893_{0.036}$ & \bR$0.967_{0.007}$ & \bR$0.857_{0.029}$ & \bR$0.962_{0.008}$ & \bR$0.959_{0.012}$  \bR&$0.931$\\
  \cmidrule(lr){1-1} \cmidrule(lr){2-3} \cmidrule(lr){4-10} \addlinespace[0.5ex]
            & &                 & {FD}                & {FA}                & {F1}                & {F2}                & {F3}                 & {F4} &  GM \\
  \cmidrule(lr){1-1} \cmidrule(lr){2-3} \cmidrule(lr){4-10} \addlinespace[0.5ex]
HELP                & 4000 &  20 & $0.910$ & $0.37$ & $0.793_{0.028}$ & $0.543_{0.036}$ & $0.413_{0.015}$ & \bR$0.799_{0.004}$   &$0.602$\\
MultiPredict        & 4000 &  20 & $0.960$ & $0.45$ & $0.756_{0.026}$ & $0.567_{0.075}$ & $0.434_{0.040}$ & $0.763_{0.011}$ & $0.627$   \\
            \textbf{\GN{}}   & \bR$800$  &  20 & \bR$0.961_{0.007}$ &  \bR$0.577_{0.079}$ & \bR$0.809_{0.019}$ & \bR$0.871_{0.024}$ & \bR$0.814_{0.046}$ & $0.734_{0.142}$ & \bR$0.784$\\
\bottomrule
\end{tabular}%
}
\end{table*}

By inputting supplementary NN encodings in our predictor (as shown in Figure~\ref{fig:model_arch}), we can better represent the relative performance of NNs, especially when only a few samples are used for predictor training.
From Table \ref{table:adj_gin_enc_study}, we see that the impact of these supplementary encodings revealed an almost universal benefit: 11 out of 12 device pools displayed improved performance. This is likely due to the fact that these encodings contextualize the few target samples with respect to the broader search space more effectively. 

\subsection{Impact of Pre-Training Samples}

Focusing on the pre-training phase, we delve into how varying sample sizes from source devices influence the Spearman rank correlation. Notably, the end-to-end performance does not consistently improve with an increasing number of samples; quite the opposite, it occasionally diminishes. This counterintuitive phenomenon can be ascribed to a situation where the model, encountering a multitude of source devices that share high correlation, ends up overfitting to the specifics of this source device set. For instance, in `Task N4', where the training set already has a relatively low average correlation between devices, degradation of predictive performance with more samples isn't observed. However, in `Task N2', which comprises solely of GPUs, this tendency becomes more clear. These findings indicate that the diversity in the training pool is important to benefit from larger source sample training sets. Merely increasing the number of samples without assuring their diversity can hinder predictor pre-training.
We perform an ablation in the appendix to select a reasonable number of pretraining samples for use with our predictor in subsequent experiments.

\subsection{Combining our Optimizations and a Comparison to Prior Work}

Table \ref{table:design_components} lists the effect of combining all of our optimizations including the hardware-aware operation embeddings, embedding initialization, encoding-based samplers, and supplementary encodings, all performed on our predictor architecture.
We call our final predictor \textbf{\GN{}}: \textbf{N}eural \textbf{A}rchitecture \textbf{S}ampler And \textbf{F}ew-Shot \textbf{Lat}ency Predictor.
On average, our first 3 optimizations bring marked improvements to the predictor performance.
We also found that using our encoding-based samplers generally reduced variance, making predictor construction more reliable.
This is further quantified in the appendix.

Finally, Table \ref{table:all_final_results} incorporates all our optimizations to deliver state-of-the-art end-to-end latency predictor performance on 11 out of 12 device sets. 
We show that, especially on challenging tasks, our optimizations improve predictor accuracy by 22.5\%    on average, and up to 87.6\% for the hardest (F3) task when compared to HELP~\cite{help}.

\comment{
To study the effectiveness of transfer of latency predictor, we first contextualize the studies in our entire paper on four device spaces across FBNet and NASBench-201 with 20 samples per target device. Table \ref{table:design_components} indicates that initializing the hardware with its closest correlated training device serves as a significant contributor in the predictors performance. Further, the operation-wise hardware embedding methodology can boost the predictor performance as well. In the context of samplers, their contributions are less evident as per Table \ref{table:design_components}. 

This is because the impact of samplers is more evident in Table \ref{table:sampler_study}, where we use only 5 samples on the target device. We also find that samplers are more useful in reducing the variance of experiments, as indicated in Table \ref{table:design_components} and further in the Appendix. The supplemental encoding helps significantly in device sets F1, F2 and F4, but not so much on F3. We find that this disparity arises from higher correlation of supplemental encoding with the test devices in F1, F2 and F4 as opposed to F3. This observation further highlights the positive impact supplemental information can have in certain use cases, as they may exhibit high input-output correlation. 

}

\begin{table*}[t]
    \caption{Performance comparison of different latency estimators combined with MetaD2A for latency-constrained NAS, on CIFAR-100 dataset with NAS-Bench-201 search space. 
    For the building time and the total NAS cost of MetaD2A+HELP, we report only time and cost during the meta-test time. The meta-training time of HELP is 25 hours, \GN{} is 25 minutes and the time to meta-train the MetaD2A is 46 GPU hours, which is conducted only once across all unseen devices. We report average over 10 trials. S indicates number of new samples on target device. } 
    \label{table:nas_result}
    \resizebox{\linewidth}{!}{%
            \begin{tabular}{clcccHcccc}
            \toprule
            \centering
            \multirow{2}{*}{Device} & \multicolumn{1}{c}{\multirow{2}{*}{Model}} &  Const & Latency & Accuracy & MACs & \multicolumn{2}{c}{Latency Model}   & Total NAS Cost & Speed Up\\
             &                                                                  & (ms)   & (ms)    &  (\%)    & (M)    &   Sample    & Building Time           & (Wall Clock) &   \\
            \cmidrule(lr){1-1} \cmidrule(lr){2-2} \cmidrule(lr){3-6} \cmidrule(lr){7-8} \cmidrule(lr){9-10} \addlinespace[0.5ex]
                                                                & MetaD2A + BRP-NAS \cite{brpnas}  & \multicolumn{1}{c}{\multirow{2}{*}{14}}         & 14 & 66.9 & 79 & 900 & 1120s & 1220s & 0.1$\times$ \\
                                                                & MetaD2A + HELP \cite{help}       &                                                 & 13 & 67.4 & 47 & 20 & 25s & 125s & 1$\times$ \\
                                                                & \textbf{MetaD2A + \GN{}}                   &                                                 & \bK{$14.4_{3.41}$} & \bK{$68.53_{3.04}$} & 47 & \textbf{20} & \textbf{25s} & \textbf{29.1s} & \textbf{4.3$\times$} \\
            \cmidrule(lr){2-2} \cmidrule(lr){3-6} \cmidrule(lr){7-8} \cmidrule(lr){9-10} \addlinespace[0.5ex]
            Unseen Device                                       & MetaD2A + BRP-NAS \cite{brpnas}  & \multicolumn{1}{c}{\multirow{2}{*}{22}}         & 34 & 73.5 & 185 & 900 & 1120s & 1220s  & 0.1$\times$ \\
            Google Pixel2                                       & MetaD2A + HELP \cite{help}       &                                                 & 19 & 70.6 & 55 & 20 & 25s & 125s & 1$\times$ \\
                                                                & \textbf{MetaD2A + \GN{}}                   &                                                 & \bK{$22.2_{6.46}$}  & \bK{$72.08_{0.91}$}  & 47 & \textbf{20} & \textbf{25s} & \textbf{29.1s} & \textbf{4.3$\times$} \\
            \cmidrule(lr){2-2} \cmidrule(lr){3-6} \cmidrule(lr){7-8} \cmidrule(lr){9-10} \addlinespace[0.5ex]
                                                                & MetaD2A + BRP-NAS \cite{brpnas}  & \multicolumn{1}{c}{\multirow{2}{*}{34}}         & 34 & 73.5 & 185 & 900 & 1120s & 1220s  & 0.1$\times$ \\
                                                                & MetaD2A + HELP \cite{help}       &                                                 & 34 & 73.5 & 185 & 20 & 25s & 125s & 1$\times$ \\
                                                                & \textbf{MetaD2A + \GN{}}                   &                                                 & \bK{$34_{0.0}$}  & \bK{$73.5_{0.0}$}     & 47 & \textbf{20} &\textbf{25s} & \textbf{29.1s} & \textbf{4.3$\times$} \\
            \cmidrule(lr){1-1} \cmidrule(lr){2-2} \cmidrule(lr){3-6} \cmidrule(lr){7-8} \cmidrule(lr){9-10} \addlinespace[0.5ex]
                                                                & MetaD2A + Layer-wise Pred.       & \multicolumn{1}{c}{\multirow{3}{*}{18}}         & 37 & 73.2 & 121 & 900 & 998s & 1098s & 0.1$\times$ \\
                                                                & MetaD2A + BRP-NAS \cite{brpnas}  &                                                 & 21 & 67.0 & 86 & 900 & 940s & 1040s & 0.1$\times$ \\  
                                                                & MetaD2A + HELP \cite{help}       &                                                 & 18 & 69.3 & 51 & 20 & 11s & 111s & 1 $\times$ \\
                                                                & \textbf{MetaD2A + \GN{}}                   &                                                 & \bK{$17.10_{2.65}$}   &  \bK{$69.92_{2.35}$}   & 47 & \textbf{20} &\textbf{11s} & \textbf{15.4s} & \textbf{7.2$\times$} \\
            \cmidrule(lr){2-2} \cmidrule(lr){3-6} \cmidrule(lr){7-8} \cmidrule(lr){9-10} \addlinespace[0.5ex]
            Unseen Device                                       & MetaD2A + Layer-wise Pred.       & \multicolumn{1}{c}{\multirow{3}{*}{21}}         & 41 & 73.5 & 184 & 900 & 998s & 1098s & 0.1$\times$ \\
            Titan RTX                                           & MetaD2A + BRP-NAS \cite{brpnas}  &                                                 & 19 & 71.5 & 55 & 900 & 940s & 1040s & 0.1$\times$ \\  
            (Batch Size 256)                                    & MetaD2A + HELP \cite{help}       &                                                 & 19 & 71.6 & 55 & 20 & 11s & 111s & 1 $\times$ \\
                                                                & \textbf{MetaD2A + \GN{}}                   &                                                 & \bK{$20.47_{2.46}$}   & \bK{$71.45_{0.45}$}     & 47 & \textbf{20} &\textbf{11s} & \textbf{15.4s} & \textbf{7.2$\times$} \\
            \cmidrule(lr){2-2} \cmidrule(lr){3-6} \cmidrule(lr){7-8} \cmidrule(lr){9-10} \addlinespace[0.5ex]
                                                                & MetaD2A + Layer-wise Pred.       & \multicolumn{1}{c}{\multirow{3}{*}{25}}         & 41 & 73.2 & 184 & 900 & 998s & 1098s & 0.1$\times$ \\
                                                                & MetaD2A + BRP-NAS \cite{brpnas}  &                                                 & 23 & 70.7 & 82 & 900 & 940s & 1040s & 0.1$\times$ \\  
                                                                & MetaD2A + HELP \cite{help}       &                                                 & 25 & 71.8 & 86 & 20 & 11s & 111s & 1 $\times$ \\
                                                                & \textbf{MetaD2A + \GN{}}                   &                                                 & \bK{$26.61_{4.84}$}   &  \bK{$71.9_{0.87}$}    & 47 & \textbf{20} &\textbf{11s} & \textbf{15.4s} & \textbf{7.2$\times$} \\
                                                                \bottomrule

            \end{tabular}%
        }
\end{table*}
\begin{figure*}
    \centering
    \includegraphics[width=\linewidth]{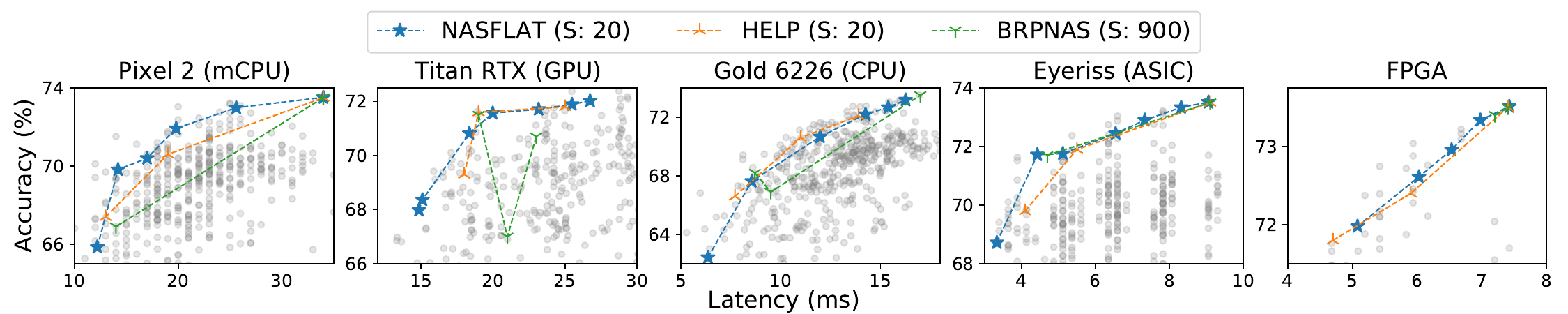}
\end{figure*}

\subsection{Neural Architecture Search}
To assess the performance on end-to-end NAS, our predictor is used within the hardware-aware NAS system presented in HELP~\cite{help}. 
We use the same NAS system with Metad2a~\cite{lee2021rapid} as the NN search algorithm for accuracy, and our predictor to find latency.
We evaluate our results using multiple metrics. Primarily, we consider the number of architecture-latency pair samples taken from the target device. 
These samples are crucial for constructing the latency predictor, which encompasses the total time span of sample acquisition, architecture compilation on the device and the latency measurement process. 
While this metric remains the same between our method and HELP, we also evaluated the total NAS cost, a combination of the time taken for few-shot transfer of predictor to the NAS experiment device, and the time spent on latency prediction during NAS. 
The results in Table~\ref{table:nas_result} (and companion plots) demonstrate a consistent improvement in the discovered NNs, with the latency-accuracy Pareto curve dominating that of prior works in most cases.
Additionally, our predictor is faster to invoke when compared to HELP, making fast NAS methods---such as Metad2a---around 5$\times$ faster.

\section{Conclusion}
In this paper, we systematically examined several design considerations inherent to hardware latency predictor design. We studied the influence of operation-specific hardware embeddings, graph neural network module designs, and the role of neural network encodings in enhancing predictor accuracy. Our exhaustive empirical evaluations across multiple device sets yielded key insights, such as the importance of sample diversity during pre-training and the significant impact of supplemental encodings based on their correlation with test devices. By adopting an objective algorithmic strategy for device set selection, we achieved a more unbiased view of latency predictor performance. Leveraging these insights, we developed and validated \GN{}, a latency predictor that outperformed existing works in 11 of 12 device sets. Future endeavors to refine predictors could involve exploring sophisticated transfer learning techniques akin to HELP \cite{help}, deepening our understanding of neural network encodings' mechanisms as samplers, and investigating various sampling methodologies.
Our work paves the way for more reliable and efficient use of predictors, both within NAS, and more generally in the optimization of NN architectures.

\bibliography{example_paper}
\bibliographystyle{mlsys2024}

\appendix

\section{Supplemental Material}

\subsection{Best practices for NAS}
\label{subsec:nasbestprac}
\citep{encodingstudy, randomnas, nasbench101, yang2019evaluation} discuss improving reproducibility and fairness in experimental comparisons for NAS. We thus address the sections released in the NAS best practices checklist by \citep{lindauer2019best}.

\begin{itemize}
    \item \textbf{Best Practice: Release Code for the Training Pipeline(s) you use: } We release code for our Predictor, CATE, Arch2Vec encoder training set-up.
    \item \textbf{Best Practice: Release Code for Your NAS Method: } We do not conduct NAS.
    \item \textbf{Best Practice: Use the Same NAS Benchmarks, not Just the Same Datasets: } We use the NASBench-201 and FBNet datasets for evaluation. We also use a sub-set of Zero Cost Proxies from NAS-Bench-Suite-Zero.
    \item \textbf{Best Practice: Run Ablation Studies: } We run extensive ablation studies in our paper. We conduct ablation studies with different supplementary encodings in the main paper as well as predictor ablations in the appendix.
    \item \textbf{Best Practice: Use the Same Evaluation Protocol for the Methods Being Compared: } We use the same evaluation protocol as existing works in this field (HELP \cite{help} and MultiPredict \cite{multipredict})
    \item \textbf{Best Practice: Evaluate Performance as a Function of Compute Resources: } In this paper, we study the sample efficiency of latency predictors. We report results in terms of the 'number of trained models required'. This directly correlates with compute resources, depending on the NAS space training procedure.
    \item \textbf{Best Practice: Compare Against Random Sampling and Random Search: } We propose a end-to-end predictor design methodology, not a NAS method.
    \item \textbf{Best Practice: Perform Multiple Runs with Different Seeds: } Our appendix contains information on number of trials and our tables in the main paper are with standard deviation.
    \item \textbf{Best Practice: Use Tabular or Surrogate Benchmarks If Possible: } All our evaluations are done on publicly available Tabular and Surrogate benchmarks.
\end{itemize}

\begin{figure*}
    \centering
    \caption{Standard deviation of neural network samplers using supplementary encodings are generally lower than random methods.}
    \label{fig:sampler_stdev}
    \includegraphics[width=\linewidth]{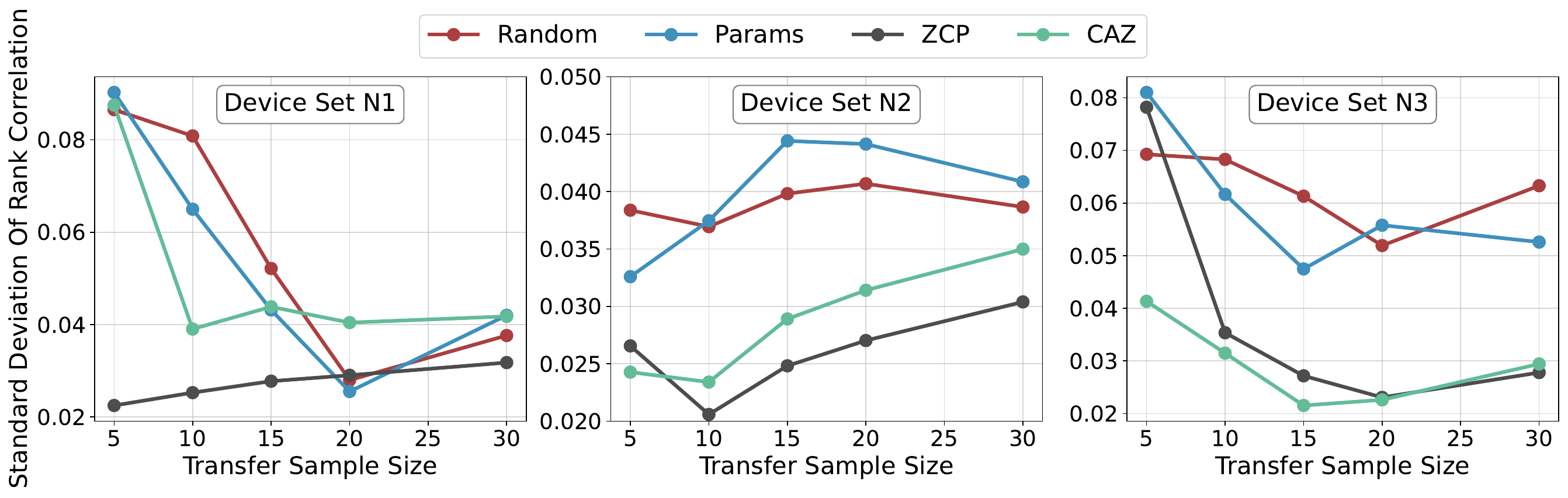}
\end{figure*}

\begin{figure*}[h!]
    \centering
    \caption{Latency-Accuracy NAS results for different sample sizes (S)}
    \label{fig:all_nas_results}
    \includegraphics[width=\linewidth]{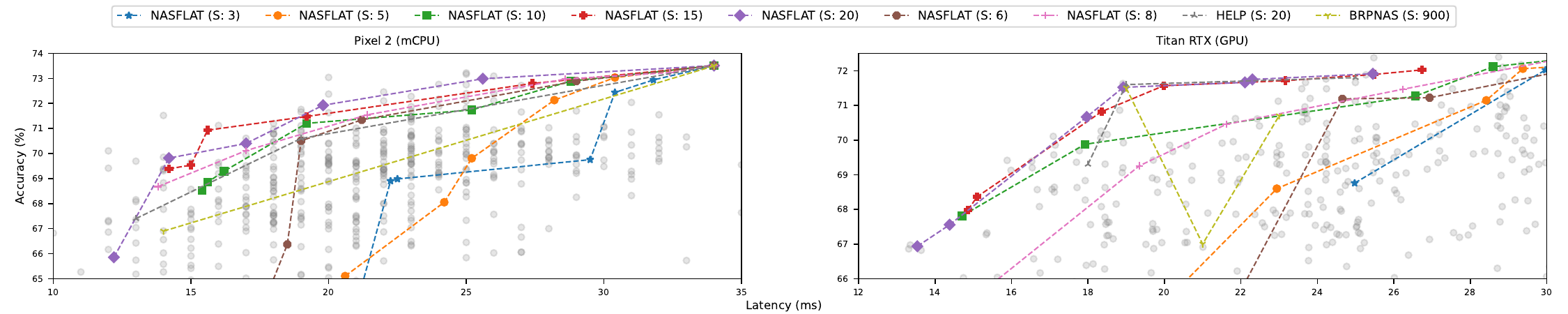}
    \includegraphics[width=\linewidth]{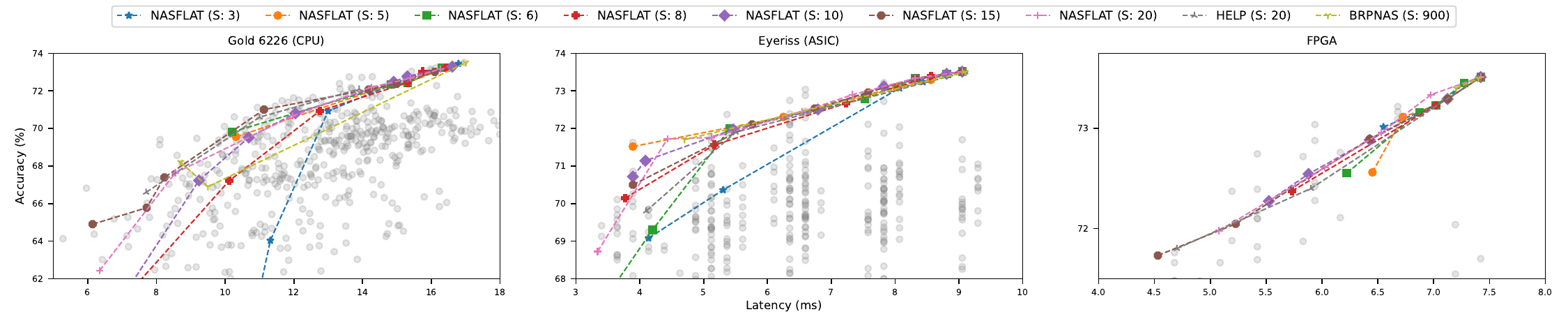}
\end{figure*}

\begin{table}[t!]
  \centering
\caption{In our tests, cosine based selection for samplers with vector encodings outperforms kMeans. Tested on Task N3 with operation-wise hardware embedding and hardware embedding initialization.}\vspace{2mm}
\label{table:kmeans_cosine}
  \resizebox{0.73\linewidth}{!}{%
\begin{tabular}{ccccc}
\toprule
& \multicolumn{4}{c}{10 Samples}         \\
\cmidrule(lr){1-1} \cmidrule(lr){2-5}\addlinespace[0.5ex]
NB201  & ZCP      & Arch2Vec & CATE     & CAZ          \\
\cmidrule(lr){1-1} \cmidrule(lr){2-5}\addlinespace[0.5ex]
Cosine & \bR$0.948$ & \bR$0.949$ & \bR$0.933$ & \bR$0.951$  \\
Kmeans & $0.729$ & $0.670$ & $0.826$ & $0.892$  \\
\cmidrule(lr){1-1} \cmidrule(lr){2-5}\addlinespace[0.5ex]
FBNet  & ZCP      & Arch2Vec & CATE     & CAZ            \\
\cmidrule(lr){1-1} \cmidrule(lr){2-5}\addlinespace[0.5ex]
Cosine & \bR$0.822$ & \bR$0.803$ & \bR$0.805$ & \bR$0.788$ \\
Kmeans & $0.412$ & $0.657$ & NaN      & $0.718$ \\
\cmidrule(lr){1-1} \cmidrule(lr){2-5}\addlinespace[0.5ex]
& \multicolumn{4}{c}{20 Samples}            \\
\cmidrule(lr){1-1} \cmidrule(lr){2-5}\addlinespace[0.5ex]
NB201  & ZCP      & Arch2Vec & CATE     & CAZ      \\
\cmidrule(lr){1-1} \cmidrule(lr){2-5}\addlinespace[0.5ex]
Cosine & \bR$0.963$  & \bR$0.960$ & \bR$0.952$ & \bR$0.960$ \\
Kmeans & $0.786$ & $0.680$  & $0.885$  & $0.948$ \\
\cmidrule(lr){1-1} \cmidrule(lr){2-5}\addlinespace[0.5ex]
FBNet  & ZCP      & Arch2Vec & CATE     & CAZ      \\
\cmidrule(lr){1-1} \cmidrule(lr){2-5}\addlinespace[0.5ex]
Cosine & \bR$0.845$ & \bR$0.839$ & \bR$0.828$ & \bR$0.852$ \\
Kmeans & $0.818$  & $0.812$ & NaN      & $0.835$  \\\bottomrule
\end{tabular}%
}
\end{table}

\subsection{Experimental settings for tables}
Table \ref{table:datasets}: Random seeds are used to generate 4 different device sets for NB201 and FBNet each.\\
Table \ref{table:opwise_emb} uses  Random sampler without supplementary encoding, 20 samples on the target device. No supplementary encoding is used.\\
Table \ref{table:sampler_study}: only 5 samples on the test device are used for transfer to effectively test different samplers under few-shot conditions. No supplementary encoding is used.\\
Table \ref{table:adj_gin_enc_study}: CAZ + kMeans was used as the sampler and 20 samples are fetched for transfer to ensure effective training of baseline predictor.\\
Table \ref{table:gcn_gat_design}:  20 samples are used for transfer with random sampler, no supplementary encoding.\\
Table \ref{table:all_final_results}: Predictor is trained  with the CAZ and CATE sampler with ZCP and Arch2Vec supplemental encoding for NASBench-201 and FBNet respectively.\\Table \ref{table:design_components}: ZCP and Arch2Vec supplemental encoding, CAZ and CATE sampler for NASBench-201 and FBNet respectively. 20 samples used for transfer to target device.\\

\subsection{NAS Search Spaces}
Latency predictors for neural architecture search (NAS) generally operate on a pre-defined search space of neural network architectures. Several such architecture search spaces can be represented as cells where nodes represent activations and edges represent operations. In this paper, we evaluate  a wide range of hardware devices across 11 representative platforms described in Table \ref{table:datasets} on two neural architecture search spaces, NASBench-201 and FBNet. The hardware latency data-set generated for our tests is collated from HW-NAS-Bench \cite{hwnasbench} and Eagle \cite{brpnas}.

\textbf{Micro Cell Space (NASBench-201)} \cite{dong2020nasbench201} is a cell-based architecture design with each cell comprising 4 nodes and 6 edges. Edges can have one of five types: zerorize, skip-connection, 1$\times$1 convolution, 3$\times$3 convolution, or 3$\times$3 average-pooling. With 15625 unique architectures, the entire network is assembled using a stem cell, three stages of five cell repetitions, followed by average-pooling and a final softmax layer.

\textbf{Macro Cell Space (FBNet)} \cite{fbnet} features a fixed macro architecture with a layer-wise search space. It offers 9 configurations of 'candidate blocks' across 22 unique positions, leading to approximately $10^{21}$ potential architectures. Despite its macro nature, FBNet can be cell-represented with 22 operational edges. For consistency, we model both NASBench-201 and FBNet using adjacency and operation matrices.

\subsubsection{GNN Module Design}
    Despite the improved performance that GCNs can deliver, they suffer from an over-smoothing problem \cite{gcn_residual_gcnii}, where this is a gradual loss of discriminative information between nodes due to the convergence of node features across multiple aggregation layers. To this end, GATES \cite{gates,tagates} introduced a custom GCN module referred to as Dense Graph Flow (DGF), which utilizes residual connections within the DGF to improve performance. Additionally, we study another node propagation mechanism based on graph attention. 
    
    \textbf{Dense Graph Flow (DGF):} The Dense Graph Flow (DGF) module implements residual connections to retain localized, discriminative features. To describe this mathematically, consider \(X^{l}\) as the input feature matrix for layer \(l\), \(A\) as the adjacency matrix, and \(O\) as the operator embedding. The corresponding parameters and bias terms for this layer are represented by \(W^{l}_{o}\), \(W^{l}_{f}\), and \(b^{l}_{f}\). Using these, the input feature matrix for the subsequent layer, \(X^{l+1}\), is determined using the sigmoid activation function, \(\sigma\), as:

\begin{equation}
X^{l+1} = \sigma(O W^{l}_{o}) \odot (A X^{l}W^{l}_{f}) + X^{l}W^{l}_{f} + b^{l}_{f}
\label{eq:dgf}
\end{equation}

\textbf{Graph Attention (GAT):}
The GAT approach \citep{gat_bengio} distinguishes itself from DGF by its attention mechanism during node information aggregation. Rather than utilizing a linear transformation \(W^{l}_{o}\) like in DGF for operation features, GAT assesses pairwise interactions among nodes via its dedicated attention layer. For layer \(l\), node features (or input feature matrix) are denoted by \(X^{l}\). A linear transformation characterized by the projection matrix \(W^{l}_{p}\) uplifts the input to advanced features. Subsequently, node features undergo self-attention via a common attentional mechanism, denoted as \(a\). \(S\) refers to SoftMax and \(L\) refers to LeakyReLU. Hence, the output \(X^{l+1}\) is formulated as:

\begin{equation}
\text{Attn}_{j}(X^{l}) = \text{S}(\text{L}(A_{j} \cdot a(W^{l}_{p}X^{l} \odot W_{p}X^{l}_{j}))) \odot W_{p}X^{l}_{j}
\label{eq:attention-coefficients}
\end{equation}

\begin{equation}
X^{l+1} = \text{LayerNorm}\left( \sigma(OW^{l}_{o}) \odot \sum_{j=1}^{n} \text{Attn}_{j}(X^{l}) \right)
\label{eq:graph-attention-output}
\end{equation}

Here, \( \text{Attn}_{j} \) represents the normalized attention coefficients, and \( \sigma \) is the sigmoid activation function. To enhance GAT's efficacy, we integrate the learned operation attention mechanism \(W_{o}\) (as mentioned in Equation \ref{eq:dgf}) with pairwise attention. This combined attention scheme fine-tunes the aggregated information. Additionally, we incorporate LayerNorm to ensure training stability.

\subsection{Predict Design Ablation}
In this subsection, we conduct an in-depth study of predictor design inspired by recent research, but from an accuracy maximization perspective. We use our ablation on accuracy to design a state-of-the-art predictor that we use for our latency study. To achieve a fair evaluation, we test each design improvement on a huge set of neural architecture design spaces detailed below.

\begin{figure*}[h!]
    \centering
    \caption{A study investigating the effect of number of latency samples per training (source) device and its impact on predictor performance. We use 20 samples from the target device.}
    \label{fig:source_samp_importance}
    \includegraphics[width=0.97\linewidth]{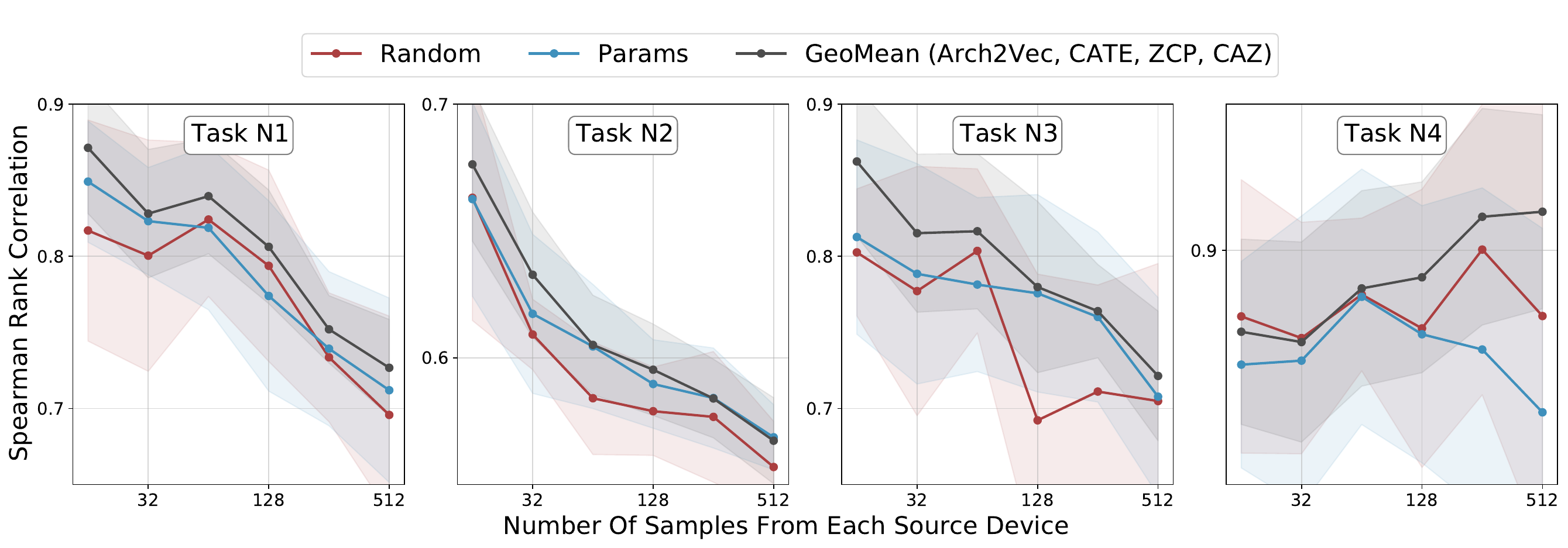}
\end{figure*}

\begin{table*}[h!]
    \centering
    \resizebox{\linewidth}{!}{%
\begin{tabular}{l|cc|cc|cc|cc|cc|cc|cc|cc}
\toprule
Space & \multicolumn{2}{c|}{Amoeba} & \multicolumn{2}{c|}{DARTS} & \multicolumn{2}{c|}{ENAS} & \multicolumn{2}{c|}{ENAS\_fix-w-d} & \multicolumn{2}{c|}{NASNet} & \multicolumn{2}{c|}{PNAS} & \multicolumn{2}{c|}{nb101} & \multicolumn{2}{c}{nb201} \\
All Node Encoding &  False &  True  &  False &  True  &  False &  True  &        False &  True  &  False &  True  &  False &  True  &  False &  True  &  False &  True  \\
Samples &        &        &        &        &        &        &              &        &        &        &        &        &        &        &        &        \\
\midrule
8   &\bR0.100&  0.078 &\bR0.079&  0.076 &  0.075 &\bR0.078&      \bR0.183&  0.154 &\bR0.135&  0.122 &  0.082 &\bR0.089&\bR0.395&  0.370 &  0.445 &\bR0.533\\
16  &\bR0.202&  0.157 &  0.178 &\bR0.184&  0.165 &\bR0.186&      \bR0.322&  0.302 &\bR0.155&  0.127 &  0.124 &\bR0.127&\bR0.448&  0.350 &  0.647 &\bR0.656\\
32  &  0.287 &\bR0.293&  0.246 &\bR0.251&\bR0.295&  0.277 &      \bR0.319&  0.301 &  0.223 &  0.223 &  0.246 &\bR0.251&\bR0.543&  0.532 &  0.709 &\bR0.715\\
64  &\bR0.375&  0.367 &\bR0.416&  0.411 &\bR0.364&  0.357 &      \bR0.374&  0.372 &\bR0.347&  0.331 &\bR0.348&  0.330 &\bR0.652&  0.623 &\bR0.771&  0.762 \\
128 &\bR0.455&  0.419 &\bR0.476&  0.474 &\bR0.463&  0.442 &        0.386 &\bR0.388&\bR0.432&  0.398 &\bR0.505&  0.494 &\bR0.703&  0.698 &\bR0.818&  0.808 \\
\bottomrule
\end{tabular}%
}
\caption{Investigating the impact of whether we should use the features of \textbf{every} node for the backward MLP (True) or not (False).}
\label{tab:allYtoMLP}
\end{table*}

\begin{figure*}[h!]
    \centering
    \begin{subfigure}
        \centering
        \includegraphics[width=0.9\columnwidth]{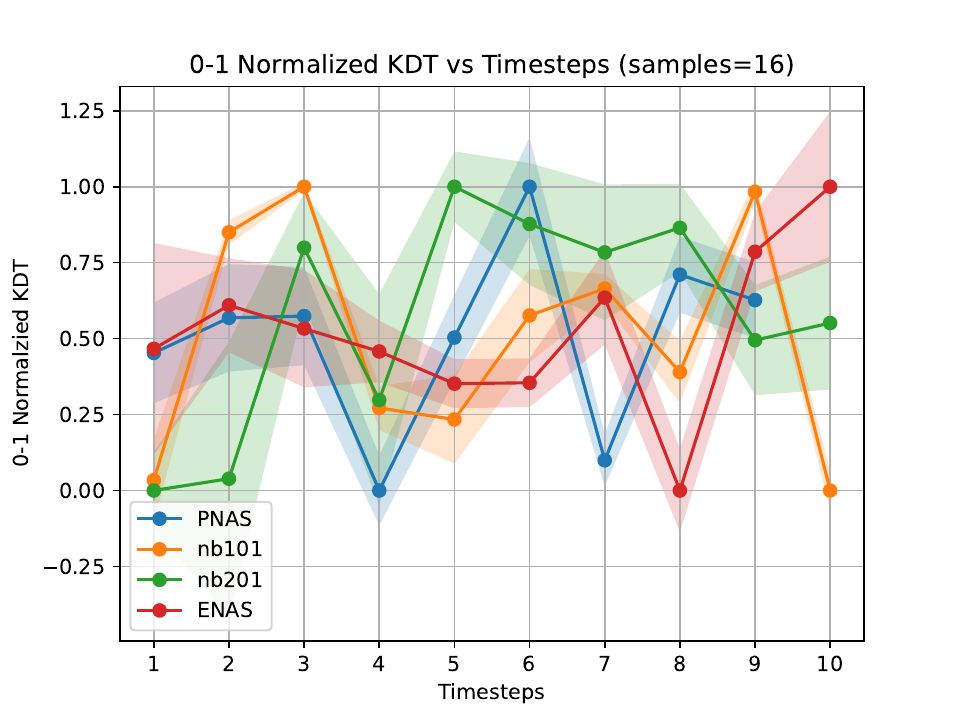}
    \end{subfigure}%
    \begin{subfigure}
        \centering
        \includegraphics[width=0.9\columnwidth]{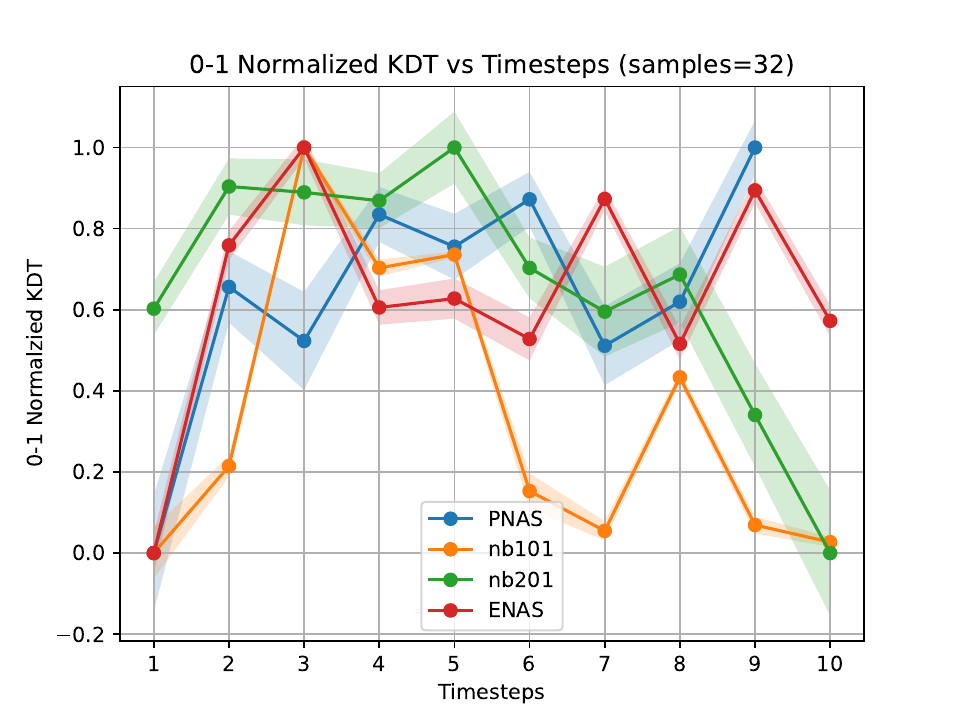}
    \end{subfigure}
    
    \begin{subfigure}
        \centering
        \includegraphics[width=0.9\columnwidth]{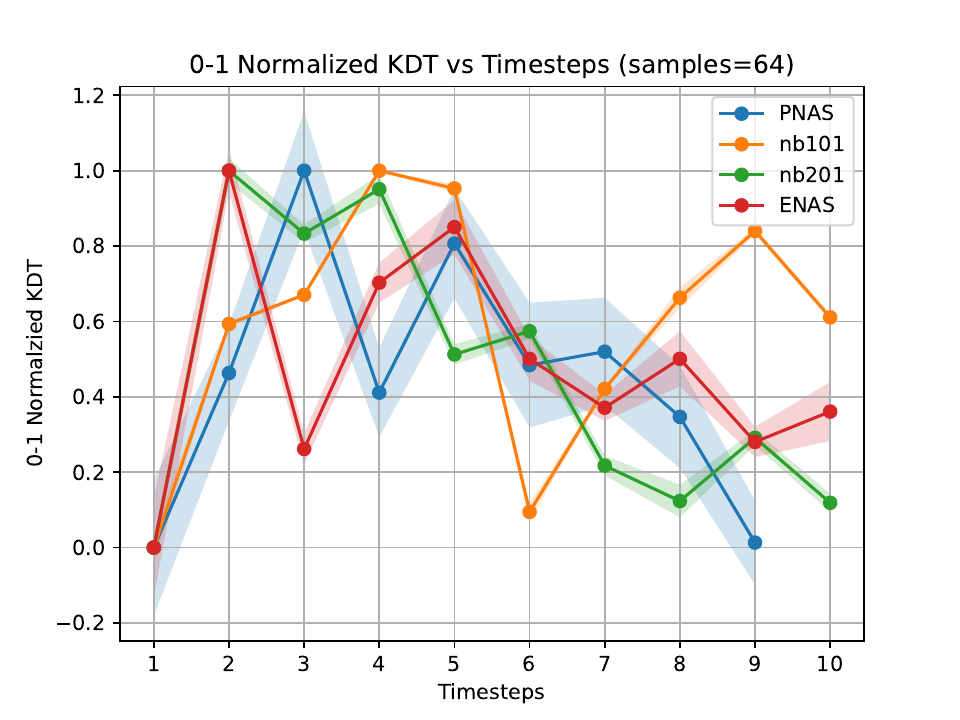}
    \end{subfigure}%
    \begin{subfigure}
        \centering
        \includegraphics[width=0.9\columnwidth]{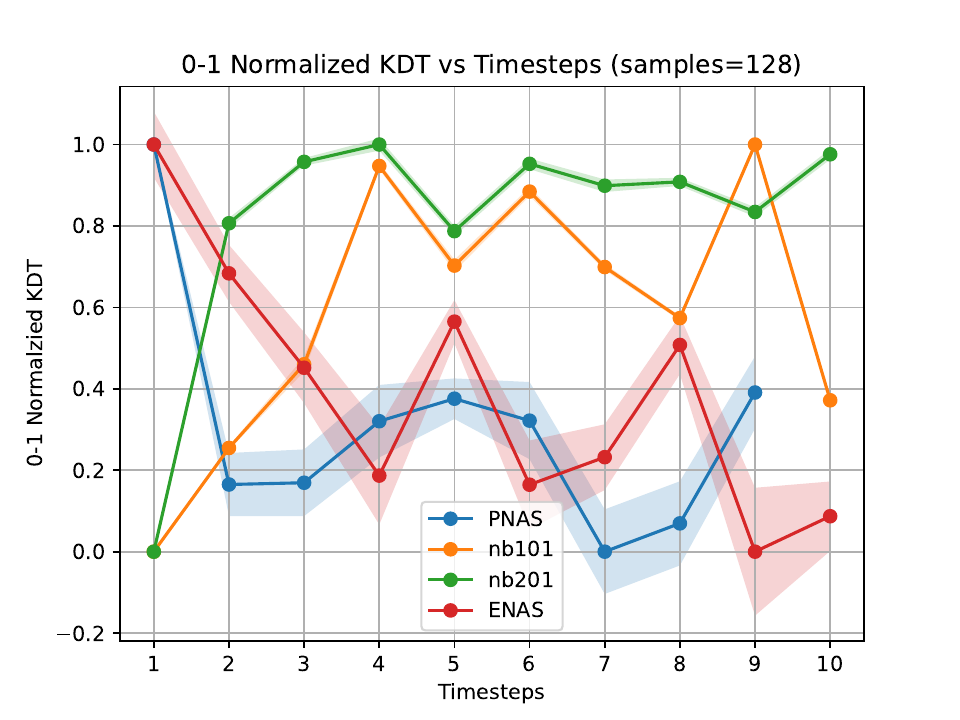}
    \end{subfigure}
    \caption{Testing the impact of time-steps in the training-analogous operation update regime. 0-1 normalized KDT for each space.}
    \label{fig:tiemstep_abl}
\end{figure*}

\subsubsection{Neural Architecture Design Spaces}
\label{subsec:nasspacesabl}

In this study, we examine various unique neural architecture design domains. We delve into NASBench-101 \citep{nasbench101} and NASBench-201 \citep{dong2020nasbench201}, both of which are cell-based search spaces encompassing 423,624 and 15,625 architectures, respectively. While NASBench-101 is trained on CIFAR-10, NASBench-201 benefits from training on CIFAR-10, CIFAR-100, and ImageNet16-120. NASBench-301 \citep{nasbench301} acts as a surrogate NAS benchmark with an impressive count of $10^{18}$ architectures. Meanwhile, TransNAS-Bench-101 \citep{transnasbench} offers both a micro (cell-based) search area featuring 4096 architectures and a broader macro search domain with 3256 designs. For the scope of our study, we focus on the TransNASBench-101 Micro due to its cell-based nature. Each of these networks undergoes training across seven distinct tasks sourced from the Taskonomy dataset. The NASLib framework brings coherence to these search areas. NAS-Bench-Suite-Zero \citep{nasbenchsuitezero} expands the landscape by introducing two datasets from NAS-Bench-360 and four more from Taskonomy. It's worth noting that the NDS dataset features `$FixWD$' datasets, signifying that the architectures maintain consistent width and depth.

\subsubsection{Training analogous predictor training}
Given a DAG, the TA-GATES\cite{tagates} encoding begins by obtaining the initial operation embedding for all operations based on their types. For $T$ time steps, an iterative process updates the operation embeddings, mimicking architecture parameter updates in training. Each step involves computing information flow via GCNs on the architecture DAG, and calling an MLP using the previous operation embedding; the output is then used in a backward GCN pass and then computing the operation embedding update. The last step concatenates the previous operation embedding with the forward and backward propagated information, feeds this to an MLP, and yields the updated operation embedding for the next step. The final architecture encoding uses the output of the $T$-th iterative refinement. In Figure \ref{fig:tiemstep_abl} we study the impact of changing the number of time-steps on the over-all kendall tau correlation (KDT). The trend is inconsistent, and therefore we attempt to investigate the utility of the backward GCN. 

\begin{table*}[t!]
  \centering
\caption{Sample Size 64 (left), 128 (right)}
  \resizebox{\linewidth}{!}{%
\begin{tabular}{cccHc}
\toprule
Space & DOpEmbUnrolled BMLP & Default & &  DOpEmbUnrolled GCN \\\cmidrule(lr){1-1}\cmidrule(lr){2-5} \addlinespace[0.5ex]
PNAS & 0.3244 & 0.3  & 0.3075  & \bR0.3395\\
ENAS & \bR0.409 & 0.3929  &  0.3873  & 0.3764\\
nb201 & \bR0.7831 & 0.7795 & 0.7455  & 0.7640 \\
nb101 & 0.6055 & \bR0.6283 & 0.5402  &  0.6041 \\

\bottomrule
\end{tabular}

\begin{tabular}{ccHc}
\toprule
 DOpEmbUnrolled BMLP & Default & &  DOpEmbUnrolled GCN \\\midrule
 \bR0.4779 & 0.4684  & 0.4588   & 0.4481\\
 0.4925 & \bR0.4958  & 0.4679   & 0.4498\\
 0.7953 & \bR0.8007  & 0.7464  & 0.7838 \\
 \bR0.7122 & 0.7013 &  0.6316   & 0.6993\\
\bottomrule
\end{tabular}%
}
\label{tab:opemb_direct} 
\end{table*}

To conduct a deeper investigation of this phenomenon, we study several aspects of training analogous predictor training. We look at \texttt{BMLP}, which is where we replace the backward 'GCN' with a small MLP. Further, the backward GCN pass uses the output of the 'forward' GCN along with the transposed adjacency matrix. This is then passed to an 'operation update MLP' that takes as input \texttt{BYI}, which is the output of the backward GCN, \texttt{BOpE}, which is the operation embedding itself. In Table \ref{tab:pnas_fb_ablation}, Table \ref{tab:enas_fb_ablation}, Table \ref{tab:nb101_fb_ablation}, Table \ref{tab:nb201_fb_ablation} we can see that in all cases, \texttt{BMLP} outperforms having a backward GCN. Additionally, in many cases the \texttt{BYI} information does not add much value. However, it does not harm performance. Therefore, for further tests we will use \texttt{BMLP} with \texttt{BYI,BOpE}.

\subsubsection{Inputs to backward MLP and gradient flow}
From Figure \ref{fig:tiemstep_abl}, we have observed that 2 timesteps generally help but more timesteps are not useful for predictor performance. Additionally, we have replaced the entire backward GCN with a backward MLP. We now investigate which gradients need to be detached during iterative refinement. In the 'def' (default) TA-GATES case, the \texttt{BYI} is \textbf{not detached}, whereas the \texttt{BOpE} is detached. In our tests, in 'all', we detach \texttt{BYI,BOpE} and in 'none', we do not detach any inputs to the \texttt{BMLP}. We find no clear pattern over 8 tests (3 trials each) in detach, except that it is better to either use the default rule or detach none of the gradients. For simplicity, we will detach none of the gradients. We again see that \texttt{BYI} is important for the \texttt{BMLP}, but the utility of \texttt{BOpE} is unclear. Further, In Table \ref{tab:allYtoMLP}, we test whether we need to pass only the encoding at the output node of the forward GCN, or should we concatenate encodings at all nodes to pass to the backward MLP/GCN. Here, we find that there is no clear advantage of passing all node encodings and thus we only use the output node encoding.

\subsubsection{Unrolled backward MLP computation}
Here, we investigate different ways to unroll the backward computation to simplify the encoding process even further. In Table \ref{tab:opemb_direct}, we introduce two methods of unrolling the computation. We only unroll for 2 time-steps, which gives us a computational graph very similar to Figure \ref{fig:model_arch}. In the case of \texttt{DOpEmbUnrolled BMLP}, (Direct Op-Emb Unrolled to BMLP) we directly take the output of the forward GNN along with the operation embedding, pass it to an MLP and use that as the encoding for the next GNN. In the case of \texttt{DOpEmbUnrolled GCN} (Direct Op-Emb Unrolled to GCN) we directly take the output of the forward GNN along with the operation embedding and pass it to the backward-GCN instead of the BMLP, and use the output as an encoding for the next GNN. We find that unrolling the computation further improves predictive performance. 

\subsubsection{Final Predictor Architecture Design}
Finally, this leads us to our own architecture design. In our architecture, we significantly simplify the predictor architecture. Firstly, we maintain a smaller GNN which refines the operation and hardware embedding. This refind embedding is passed to an MLP which maps the embedding back to the original dimensions. This refined embedding is passed directly to the larger GNN along with the adjacency matrix and node embeddings. We find that this simplified architecture performs better in most of our tests.

\begin{table*}[t!]
  \centering
\caption{ENAS Sample Size 64 (left), 128 (right). Ablation for backward pass. BMLP indicates that instead of replicating network, we do a simple 2 layer MLP for backward pass. BYI indicates whether we use the output of the forward pass network or not. BOpE indicates whether we use the output of the operation embedding itself or not. PM indicates that num params are approximately matched wrt TS $>$ 1. w / d indicates whether the matching happens by adjusting width or depth in forward gcn architecture. 2R implies that we simply use a small random perturb vector for operation update to 'regularize' the network. }
  \resizebox{0.8\linewidth}{!}{%
\begin{tabular}{|c|c|c|cH|c|c|}
\toprule
TS & BMLP & BYI & BOpE & O-BOpE & KDT & Dev \\\midrule
2R  & \ding{55} & \ding{55} & \ding{55} & \ding{55} & 0.3397 & 0.0018 \\
1$_{PM}^{d}$ & \ding{55} & \ding{55} & \ding{55} & \ding{55} & 0.3832 & 0.0027 \\
2 & \ding{51} & \ding{51} & \ding{55} & \ding{55} & 0.3941 & 0.0061 \\
3 & \ding{51} & \ding{55} & \ding{51} & \ding{55} & 0.3973 & 0.0008 \\
1$_{PM}^{w}$ & \ding{55} & \ding{55} & \ding{55} & \ding{55} & 0.3983 & 0.006 \\
3 & \ding{51} & \ding{51} & \ding{55} & \ding{55} & 0.3988 & 0.0054 \\\midrule
2 & \ding{51} & \ding{55} & \ding{51} & \ding{55} & 0.4142 & 0.0009 \\
\bottomrule
\end{tabular}

\begin{tabular}{|c|c|c|cH|c|c|}
\toprule
TS & BMLP & BYI & BOpE & O-BOpE & KDT & Dev \\\midrule
2R  & \ding{55} & \ding{55} & \ding{55} & \ding{55} & 0.4611 & 0.0010 \\
1$_{PM}^{w}$ & \ding{55} & \ding{55} & \ding{55} & \ding{55} & 0.4635 & 0.0027 \\
1$_{PM}^{d}$ & \ding{55} & \ding{55} & \ding{55} & \ding{55} & 0.4684 & 0.0007 \\
2 & \ding{51} & \ding{55} & \ding{55} & \ding{55} & 0.4686 & 0.0005 \\\midrule
3 & \ding{55} & \ding{51} & \ding{55} & \ding{55} & 0.4847 & 0.0007 \\
3 & \ding{51} & \ding{55} & \ding{51} & \ding{55} & 0.4888 & 0.0039 \\
2 & \ding{51} & \ding{55} & \ding{51} & \ding{55} & 0.4975 & 0.0021 \\
\bottomrule
\end{tabular}%
}
\label{tab:enas_fb_ablation} 
  \centering
\caption{NB201 Sample Size 64 (left), 128 (right). Ablation for backward pass.}
  \resizebox{0.8\linewidth}{!}{%
\begin{tabular}{|c|c|c|cH|c|c|}
\toprule
TS & BMLP & BYI & BOpE & O-BOpE & KDT & Dev \\\midrule
1$_{PM}^{d}$ & \ding{55} & \ding{55} & \ding{55} & \ding{55} & 0.7472 & 0.0005 \\
1$_{PM}^{w}$ & \ding{55} & \ding{55} & \ding{55} & \ding{55} & 0.7475 & 0.0005 \\
2R  & \ding{55} & \ding{55} & \ding{55} & \ding{55} & 0.7521 & 0.0002 \\
3 & \ding{51} & \ding{55} & \ding{55} & \ding{55} & 0.7835 & 0.0002 \\
2 & \ding{51} & \ding{51} & \ding{51} & \ding{55} & 0.7845 & 0.0001 \\
3 & \ding{51} & \ding{55} & \ding{51} & \ding{55} & 0.7856 & 0.0001 \\\midrule
3 & \ding{51} & \ding{51} & \ding{51} & \ding{55} & 0.7882 & 0.0003 \\
2 & \ding{51} & \ding{55} & \ding{51} & \ding{55} & 0.7894 & 0.0002 \\
\bottomrule
\end{tabular}

\begin{tabular}{|c|c|c|cH|c|c|}
\toprule
TS & BMLP & BYI & BOpE & O-BOpE & KDT & Dev \\\midrule
1$_{PM}^{w}$ & \ding{55} & \ding{55} & \ding{55} & \ding{55} & 0.7682 & 0.0 \\
2R  & \ding{55} & \ding{55} & \ding{55} & \ding{55} & 0.7718 & 0.0001 \\
1$_{PM}^{d}$ & \ding{55} & \ding{55} & \ding{55} & \ding{55} & 0.7765 & 0.0005 \\
2 & \ding{51} & \ding{55} & \ding{51} & \ding{55} & 0.7918 & 0.0003 \\
2 & \ding{51} & \ding{51} & \ding{51} & \ding{55} & 0.7927 & 0.0004 \\\midrule
3 & \ding{51} & \ding{51} & \ding{51} & \ding{55} & 0.7953 & 0.0005 \\
2 & \ding{51} & \ding{51} & \ding{55} & \ding{55} & 0.7956 & 0.0001 \\
3 & \ding{51} & \ding{55} & \ding{51} & \ding{55} & 0.7971 & 0.0004 \\
\bottomrule
\end{tabular}%
}
\label{tab:nb201_fb_ablation} 
  \centering
\caption{PNAS Sample Size 64 (left), 128 (right). Ablation for backward pass.}
  \resizebox{0.8\linewidth}{!}{%
\begin{tabular}{|c|c|c|cH|c|c|}
\toprule
TS & BMLP & BYI & BOpE & O-BOpE & KDT & Dev \\\midrule
1$_{PM}^{d}$ & \ding{55} & \ding{55} & \ding{55} & \ding{55} & 0.3387 & 0.008 \\
2R  & \ding{55} & \ding{55} & \ding{55} & \ding{55} & 0.3494 & 0.0165 \\
1$_{PM}^{w}$ & \ding{55} & \ding{55} & \ding{55} & \ding{55} & 0.3635 & 0.011 \\
2 & \ding{55} & \ding{51} & \ding{55} & \ding{55} & 0.364 & 0.0145 \\\midrule
2 & \ding{51} & \ding{55} & \ding{51} & \ding{55} & 0.3641 & 0.0055 \\
3 & \ding{55} & \ding{51} & \ding{55} & \ding{55} & 0.378 & 0.0106 \\
2 & \ding{51} & \ding{51} & \ding{55} & \ding{55} & 0.382 & 0.0092 \\
3 & \ding{51} & \ding{51} & \ding{55} & \ding{55} & 0.3852 & 0.0104 \\
\bottomrule
\end{tabular}

\begin{tabular}{|c|c|c|cH|c|c|}
\toprule
TS & BMLP & BYI & BOpE & O-BOpE & KDT & Dev \\\midrule
1$_{PM}^{d}$ & \ding{55} & \ding{55} & \ding{55} & \ding{55} & 0.4352 & 0.012 \\
2R  & \ding{55} & \ding{55} & \ding{55} & \ding{55} & 0.4507 & 0.0063 \\
2 & \ding{51} & \ding{55} & \ding{55} & \ding{55} & 0.4625 & 0.0035 \\\midrule
2 & \ding{51} & \ding{51} & \ding{55} & \ding{55} & 0.4684 & 0.0033 \\
3 & \ding{51} & \ding{51} & \ding{55} & \ding{55} & 0.4709 & 0.0041 \\
1$_{PM}^{w}$ & \ding{55} & \ding{55} & \ding{55} & \ding{55} & 0.4779 & 0.003 \\
3 & \ding{55} & \ding{51} & \ding{55} & \ding{55} & 0.4841 & 0.0005 \\
2 & \ding{55} & \ding{51} & \ding{55} & \ding{55} & 0.4897 & 0.0006 \\
\bottomrule
\end{tabular}%
}
\label{tab:pnas_fb_ablation} 
  \centering
\caption{NB101 Sample Size 64 (left), 128 (right). Ablation for backward pass.}
  \resizebox{0.8\linewidth}{!}{%
\begin{tabular}{|c|c|c|cH|c|c|}
\toprule
TS & BMLP & BYI & BOpE & O-BOpE & KDT & Dev \\\midrule
1$_{PM}^{d}$ & \ding{55} & \ding{55} & \ding{55} & \ding{55} & 0.6211 & 0.0007 \\
1$_{PM}^{w}$ & \ding{55} & \ding{55} & \ding{55} & \ding{55} & 0.6273 & 0.0003 \\
2 & \ding{51} & \ding{55} & \ding{55} & \ding{55} & 0.6346 & 0.0001 \\
2R  & \ding{55} & \ding{55} & \ding{55} & \ding{55} & 0.6421 & 0.0063 \\
2 & \ding{51} & \ding{51} & \ding{55} & \ding{55} & 0.6466 & 0.0008 \\
2 & \ding{51} & \ding{55} & \ding{51} & \ding{55} & 0.6502 & 0.0004 \\\midrule
3 & \ding{51} & \ding{51} & \ding{55} & \ding{55} & 0.6515 & 0.0006 \\
2 & \ding{51} & \ding{51} & \ding{51} & \ding{55} & 0.6537 & 0.0003 \\
\bottomrule
\end{tabular}

\begin{tabular}{|c|c|c|cH|c|c|}
\toprule
TS & BMLP & BYI & BOpE & O-BOpE & KDT & Dev \\\midrule
1$_{PM}^{w}$ & \ding{55} & \ding{55} & \ding{55} & \ding{55} & 0.6591 & 0.0003 \\
2R  & \ding{55} & \ding{55} & \ding{55} & \ding{55} & 0.6886 & 0.0063 \\
1$_{PM}^{d}$ & \ding{55} & \ding{55} & \ding{55} & \ding{55} & 0.7008 & 0.0001 \\
2 & \ding{51} & \ding{51} & \ding{55} & \ding{55} & 0.707 & 0.0004 \\
2 & \ding{51} & \ding{51} & \ding{51} & \ding{55} & 0.7075 & 0.0001 \\
2 & \ding{51} & \ding{55} & \ding{55} & \ding{55} & 0.7089 & 0.0002 \\\midrule
2 & \ding{51} & \ding{55} & \ding{51} & \ding{55} & 0.7194 & 0.0001 \\
3 & \ding{51} & \ding{51} & \ding{55} & \ding{55} & 0.7276 & 0.0002 \\
\bottomrule
\end{tabular}%
}
\label{tab:nb101_fb_ablation} 
\end{table*}

\begin{table*}[t!]
  \centering
\caption{ENAS Sample Size 64 (left), 128 (right). Ablation for backward pass. BMLP is always True. BYI indicates whether we use the output of the forward pass network or not. BOpE indicates whether we use the output of the operation embedding itself or not. DM indicates detachment mode. 2 timesteps fixed. }
  \resizebox{0.7\linewidth}{!}{%
\begin{tabular}{|c|c|c|c|c|}
\toprule
BYI & BOpE & DM & KDT & STD \\
\midrule
\ding{51} & \ding{55} & def & 0.4012 & 0.0062 \\
\ding{51} & \ding{55} & none & 0.4042 & 0.0044 \\
\ding{55} & \ding{51} & none & 0.4058 & 0.0013 \\
\ding{51} & \ding{55} & all & 0.4074 & 0.0053 \\
\ding{55} & \ding{51} & def & 0.4142 & 0.0009 \\
\bottomrule
\end{tabular}

\begin{tabular}{|c|c|c|c|c|}
\toprule
BYI & BOpE & DM & KDT & STD \\
\midrule
\ding{51} & \ding{51} & none & 0.4694 & 0.0025 \\
\ding{55} & \ding{55} & none & 0.4716 & 0.0005 \\
\ding{55} & \ding{51} & all & 0.4958 & 0.003 \\
\ding{55} & \ding{51} & def & 0.4975 & 0.0021 \\
\ding{55} & \ding{51} & none & 0.511 & 0.0018 \\
\bottomrule
\end{tabular}%
}
\label{tab:enas_detach_ablation} 
  \centering
\caption{NB201 Sample Size 64 (left), 128 (right). Ablation for backward pass.}
  \resizebox{0.7\linewidth}{!}{%
\begin{tabular}{|c|c|c|c|c|}
\toprule
BYI & BOpE & DM & KDT & STD \\
\midrule
\ding{55} & \ding{51} & def & 0.7795 & 0.0001 \\
\ding{51} & \ding{55} & none & 0.7853 & 0.0002 \\
\ding{51} & \ding{51} & def & 0.7859 & 0.0003 \\
\ding{55} & \ding{51} & none & 0.7871 & 0.0002 \\
\ding{51} & \ding{51} & none & 0.7949 & 0.0004 \\
\bottomrule
\end{tabular}

\begin{tabular}{|c|c|c|c|c|}
\toprule
BYI & BOpE & DM & KDT & STD \\
\midrule
\ding{51} & \ding{51} & def & 0.7875 & 0.0003 \\
\ding{51} & \ding{55} & none & 0.7888 & 0.0002 \\
\ding{51} & \ding{51} & none & 0.7936 & 0.0005 \\
\ding{55} & \ding{51} & none & 0.7944 & 0.0002 \\
\ding{55} & \ding{51} & def & 0.8007 & 0.0002 \\
\bottomrule
\end{tabular}%
}
\label{tab:nb201_detach_ablation} 
  \centering
\caption{PNAS Sample Size 64 (left), 128 (right). Ablation for backward pass.}
  \resizebox{0.7\linewidth}{!}{%
\begin{tabular}{|c|c|c|c|c|}
\toprule
BYI & BOpE & DM & KDT & STD \\
\midrule
\ding{51} & \ding{51} & def & 0.3184 & 0.0011 \\
\ding{51} & \ding{55} & none & 0.3304 & 0.0067 \\
\ding{51} & \ding{55} & all & 0.333 & 0.006 \\
\ding{51} & \ding{55} & def & 0.3459 & 0.0074 \\
\ding{51} & \ding{51} & none & 0.3538 & 0.0045 \\
\bottomrule
\end{tabular}

\begin{tabular}{|c|c|c|c|c|}
\toprule
BYI & BOpE & DM & KDT & STD \\
\midrule
\ding{51} & \ding{55} & none & 0.4388 & 0.0037 \\
\ding{51} & \ding{55} & def & 0.456 & 0.0036 \\
\ding{55} & \ding{51} & none & 0.4675 & 0.0039 \\
\ding{55} & \ding{51} & all & 0.4684 & 0.0042 \\
\ding{55} & \ding{51} & def & 0.474 & 0.0042 \\
\bottomrule
\end{tabular}%
}
\label{tab:pnas_detach_ablation} 
  \centering
\caption{NB101 Sample Size 64 (left), 128 (right). Ablation for backward pass.}
  \resizebox{0.7\linewidth}{!}{%
\begin{tabular}{|c|c|c|c|c|}
\toprule
BYI & BOpE & DM & KDT & STD \\
\midrule
\ding{51} & \ding{51} & none & 0.6329 & 0.0006 \\
\ding{51} & \ding{51} & all & 0.6432 & 0.0014 \\
\ding{55} & \ding{51} & none & 0.6436 & 0.0013 \\
\ding{55} & \ding{55} & all & 0.6552 & 0.0 \\
\ding{51} & \ding{51} & def & 0.6588 & 0.0 \\
\bottomrule
\end{tabular}

\begin{tabular}{|c|c|c|c|c|}
\toprule
BYI & BOpE & DM & KDT & STD \\
\midrule
\ding{55} & \ding{55} & def & 0.7139 & 0.0001 \\
\ding{51} & \ding{51} & all & 0.7177 & 0.0001 \\
\ding{55} & \ding{55} & none & 0.7206 & 0.0002 \\
\ding{51} & \ding{55} & def & 0.728 & 0.0 \\
\ding{51} & \ding{55} & none & 0.735 & 0.0 \\
\bottomrule
\end{tabular}%
}
\label{tab:nb101_detach_ablation} 
\end{table*}

\begin{table*}[h!]
    \centering
    \resizebox{\linewidth}{!}{%
\begin{tabular}{l|l|l|l}
\toprule
\textbf{Hyperparameter} & \textbf{Value} & \textbf{Hyperparameter} & \textbf{Value} \\ \midrule
Learning Rate & 0.001 & Weight Decay & 0.00001 \\
Number of Epochs & 150 & Batch Size & 16 \\
Number of Transfer Epochs & 40 NB201, 30 FBNet & Transfer Learning Rate & 0.003 NB201, 0.001 FBNet \\
Graph Type & \texttt{DGF+GAT ensemble} & Op Embedding Dim & 48 \\
Node Embedding Dim & 48 & Hidden Dim & 96 \\
Op-HW GCN Dims & [128, 128] & Op-HW MLP Dims & [128] \\
GCN Dims & [128, 128, 128] & MLP Dims & [200, 200, 200] \\
Number of Trials & 3 & Loss Type & Pairwise Hinge Loss \citep{tagates} \\
\bottomrule
\end{tabular}%
}
\caption{Hyperparameters used in the experiments. We run Optuna hyper-parameter optimization for 80 iterations.}
\label{tab:hyperparameters}
\end{table*}

\newpage

\begin{table*}[h!]
    \centering
    \resizebox{0.8\linewidth}{!}{%

\begin{tabular}{cccccccccc}\toprule
\multicolumn{10}{c}{NASBench201 ND Train-Test Correlation Latency Correlation between Test and Train devices} \\ \hline
{} & 1080ti\_1 & 1080ti\_32 & 1080ti\_256 & silver\_4114 & silver\_4210r & samsung\_a50 & pixel3 & essential\_ph\_1 & samsung\_s7 \\
\midrule
titan\_rtx\_256 &     0.772 &      0.792 &       0.812 &        0.947 &         0.982 &        0.975 &  0.878 &            0.897 &       0.854 \\
gold\_6226      &     0.958 &      0.956 &       0.776 &        0.912 &         0.927 &        0.894 &  0.711 &            0.898 &       0.920 \\
fpga            &     0.828 &      0.841 &       0.888 &        0.943 &         0.974 &        0.959 &  0.872 &            0.924 &       0.888 \\
pixel2          &     0.807 &      0.817 &       0.777 &        0.873 &         0.894 &        0.874 &  0.761 &            0.856 &       0.832 \\
raspi4          &     0.654 &      0.669 &       0.735 &        0.844 &         0.878 &        0.875 &  0.967 &            0.808 &       0.758 \\
eyeriss         &     0.415 &      0.434 &       0.893 &        0.586 &         0.618 &        0.625 &  0.722 &            0.624 &       0.521 \\
\bottomrule
\end{tabular}

}
\end{table*}

\begin{table*}[h!]
    \centering
    \resizebox{0.75\linewidth}{!}{%

\begin{tabular}{cccccc}\toprule
\multicolumn{6}{c}{NASBench201 N1 Train-Test Correlation Latency Correlation between Test and Train devices} \\ \hline
{} & e\_tpu\_edge\_tpu\_int8 & eyeriss & m\_gpu\_sd\_675\_AD\_612\_int8 & m\_gpu\_sd\_855\_AD\_640\_int8 & pixel3 \\
\midrule
1080ti\_1      &                   0.167 &   0.415 &                              0.594 &                              0.551 &  0.591 \\
titan\_rtx\_32 &                   0.127 &   0.403 &                              0.595 &                              0.547 &  0.599 \\
titanxp\_1     &                   0.163 &   0.405 &                              0.594 &                              0.551 &  0.589 \\
2080ti\_32     &                   0.174 &   0.424 &                              0.603 &                              0.560 &  0.601 \\
titan\_rtx\_1  &                   0.113 &   0.362 &                              0.554 &                              0.504 &  0.547 \\
\bottomrule
\end{tabular}

}
\end{table*}

\begin{table*}[h!]
    \centering
    \resizebox{0.6\linewidth}{!}{%
\begin{tabular}{cccccc}\toprule
\multicolumn{6}{c}{NASBench201 N2 Train-Test Correlation Latency Correlation between Test and Train devices} \\ \hline
{} & 1080ti\_1 & 1080ti\_32 & titanx\_32 & titanxp\_1 & titanxp\_32 \\
\midrule
e\_gpu\_jetson\_nano\_fp16          &     0.514 &      0.509 &      0.517 &      0.510 &       0.513 \\
e\_tpu\_edge\_tpu\_int8             &     0.167 &      0.172 &      0.171 &      0.163 &       0.170 \\
m\_dsp\_sd\_675\_HG\_685\_int8 &     0.593 &      0.594 &      0.599 &      0.591 &       0.596 \\
m\_dsp\_sd\_855\_HG\_690\_int8 &     0.587 &      0.583 &      0.592 &      0.585 &       0.589 \\
pixel3                              &     0.591 &      0.611 &      0.598 &      0.589 &       0.607 \\
\bottomrule
\end{tabular}

}
\end{table*}

\begin{table*}[h!]
    \centering
    \resizebox{0.8\linewidth}{!}{%
\begin{tabular}{cccccc}\toprule
\multicolumn{6}{c}{NASBench201 N3 Train-Test Correlation Latency Correlation between Test and Train devices} \\ \hline
{} & d\_gpu\_gtx\_1080ti\_fp32 & e\_gpu\_jetson\_nano\_fp16 & eyeriss & m\_dsp\_sd\_675\_HG\_685\_int8 & m\_gpu\_sd\_855\_AD\_640\_int8 \\
\midrule
1080ti\_1   &                     0.362 &                      0.514 &   0.415 &                               0.593 &                              0.551 \\
2080ti\_1   &                     0.356 &                      0.512 &   0.405 &                               0.586 &                              0.538 \\
titanxp\_1  &                     0.356 &                      0.510 &   0.405 &                               0.591 &                              0.551 \\
2080ti\_32  &                     0.371 &                      0.519 &   0.424 &                               0.598 &                              0.560 \\
titanxp\_32 &                     0.370 &                      0.513 &   0.423 &                               0.596 &                              0.564 \\
\bottomrule
\end{tabular}

}
\end{table*}

\begin{table*}[h!]
    \centering
    \resizebox{\linewidth}{!}{%
\begin{tabular}{ccccccccccc}\toprule
\multicolumn{11}{c}{NASBench201 N4 Train-Test Correlation Latency Correlation between Test and Train devices} \\ \hline
{} & d\_cpu\_i7\_7820x\_fp32 & e\_gpu\_jetson\_nano\_fp32 & e\_tpu\_edge\_int8 & eyeriss & m\_cpuSD\_855\_kryo\_485i8 & m\_dspSD\_675\_HG\_685i8 & m\_dspSD\_855\_HG\_690i8 & m\_gpuSD\_675\_AD\_612i8 & m\_gpuSD\_855\_AD\_640i8 & pixel2 \\
\midrule
1080ti\_1     &                         0.360 &                      0.739 &                   0.167 &   0.415 &                            0.645 &                               0.593 &                               0.587 &                              0.594 &                              0.551 &  0.807 \\
2080ti\_1     &                         0.353 &                      0.730 &                   0.168 &   0.405 &                            0.635 &                               0.586 &                               0.581 &                              0.581 &                              0.538 &  0.791 \\
titan\_rtx\_1 &                         0.313 &                      0.703 &                   0.113 &   0.362 &                            0.600 &                               0.547 &                               0.541 &                              0.554 &                              0.504 &  0.775 \\
\bottomrule
\end{tabular}

}
\end{table*}

\begin{table*}[h!]
    \centering
    \resizebox{\linewidth}{!}{%
\begin{tabular}{cccccccccccccccccc}\toprule
\multicolumn{18}{c}{NASBench201 NA Train-Test Correlation Latency Correlation between Test and Train devices} \\ \hline
{} & titan\_rtx\_1 & titan\_rtx\_32 & titanxp\_1 & 2080ti\_1 & titanx\_1 & 1080ti\_1 & titanx\_32 & titanxp\_32 & 2080ti\_32 & 1080ti\_32 & gold\_6226 & samsung\_s7 & silver\_4114 & gold\_6240 & silver\_4210r & samsung\_a50 & pixel2 \\
\midrule
eyeriss                   &         0.362 &          0.403 &      0.405 &     0.405 &     0.409 &     0.415 &      0.418 &       0.423 &      0.424 &      0.434 &      0.503 &       0.521 &        0.586 &      0.609 &         0.618 &        0.625 &  0.609 \\
d\_gpu\_gtx\_1080ti\_fp32 &         0.315 &          0.346 &      0.356 &     0.356 &     0.362 &     0.362 &      0.369 &       0.370 &      0.371 &      0.376 &      0.438 &       0.450 &        0.488 &      0.507 &         0.511 &        0.513 &  0.501 \\
e\_tpu\_edge\_tpu\_int8   &         0.113 &          0.127 &      0.163 &     0.168 &     0.166 &     0.167 &      0.171 &       0.170 &      0.174 &      0.172 &      0.221 &       0.246 &        0.243 &      0.268 &         0.256 &        0.261 &  0.299 \\
\bottomrule
\end{tabular}

}
\caption{NASBench-201 task train-test correlations. HG: Hexagon; AD: Adreno; m: Mobile; SD: Snapdragon; e: Embedded; i8: int8. Rows are test devices, Column headers are training devices.}
\label{tab:nb2traintestcorr}
\end{table*}

  \cleardoublepage

\begin{table*}[h!]
    \centering
    \resizebox{0.8\linewidth}{!}{%

\begin{tabular}{cccccccccc}\toprule
\multicolumn{10}{c}{FBNet FD Train-Test Correlation Latency Correlation between Test and Train devices} \\ \hline
{} & 1080ti\_1 & 1080ti\_32 & 1080ti\_64 & silver\_4114 & silver\_4210r & samsung\_a50 & pixel3 & essential\_ph\_1 & samsung\_s7 \\
\midrule
fpga    &     0.226 &      0.419 &      0.605 &        0.674 &         0.679 &        0.865 &  0.906 &            0.700 &       0.713 \\
raspi4  &     0.256 &      0.524 &      0.719 &        0.641 &         0.649 &        0.841 &  0.957 &            0.660 &       0.678 \\
eyeriss &     0.247 &      0.527 &      0.757 &        0.624 &         0.633 &        0.864 &  0.976 &            0.678 &       0.690 \\
\bottomrule
\end{tabular}

}
\end{table*}

\begin{table*}[h!]
    \centering
    \resizebox{0.5\linewidth}{!}{%
\begin{tabular}{cccccc}\toprule
\multicolumn{6}{c}{FBNet F1 Train-Test Correlation Latency Correlation between Test and Train devices} \\ \hline
{} & 2080ti\_1 & essential\_ph\_1 & silver\_4114 & titan\_rtx\_1 & titan\_rtx\_32 \\
\midrule
eyeriss      &     0.238 &            0.678 &        0.624 &         0.249 &          0.442 \\
fpga         &     0.206 &            0.700 &        0.674 &         0.217 &          0.350 \\
raspi4       &     0.241 &            0.660 &        0.641 &         0.251 &          0.450 \\
samsung\_a50 &     0.310 &            0.646 &        0.686 &         0.317 &          0.429 \\
samsung\_s7  &     0.293 &            0.650 &        0.649 &         0.312 &          0.352 \\
\bottomrule
\end{tabular}

}
\end{table*}

\begin{table*}[h!]
    \centering
    \resizebox{0.5\linewidth}{!}{%
\begin{tabular}{cccccc}\toprule
\multicolumn{6}{c}{FBNet F2 Train-Test Correlation Latency Correlation between Test and Train devices} \\ \hline
{} & essential\_ph\_1 & gold\_6226 & gold\_6240 & pixel3 & raspi4 \\
\midrule
1080ti\_1     &            0.258 &      0.207 &      0.536 &  0.249 &  0.256 \\
1080ti\_32    &            0.338 &      0.241 &      0.459 &  0.555 &  0.524 \\
2080ti\_32    &            0.312 &      0.214 &      0.449 &  0.519 &  0.492 \\
titan\_rtx\_1 &            0.268 &      0.184 &      0.536 &  0.253 &  0.251 \\
titanxp\_1    &            0.286 &      0.222 &      0.568 &  0.270 &  0.276 \\
\bottomrule
\end{tabular}

}
\end{table*}

\begin{table*}[h!]
    \centering
    \resizebox{0.5\linewidth}{!}{%
\begin{tabular}{cccccc}\toprule
\multicolumn{6}{c}{FBNet F3 Train-Test Correlation Latency Correlation between Test and Train devices} \\ \hline
{} & essential\_ph\_1 & pixel2 & pixel3 & raspi4 & samsung\_s7 \\
\midrule
1080ti\_1      &            0.258 &  0.300 &  0.249 &  0.256 &       0.307 \\
1080ti\_32     &            0.338 &  0.409 &  0.555 &  0.524 &       0.372 \\
2080ti\_1      &            0.240 &  0.287 &  0.243 &  0.241 &       0.293 \\
titan\_rtx\_1  &            0.268 &  0.296 &  0.253 &  0.251 &       0.312 \\
titan\_rtx\_32 &            0.313 &  0.369 &  0.471 &  0.450 &       0.352 \\
\bottomrule
\end{tabular}

}
\end{table*}

\begin{table*}[h!]
    \centering
    \resizebox{0.8\linewidth}{!}{%
\begin{tabular}{ccccccccccc}\toprule
\multicolumn{11}{c}{FBNet F4 Train-Test Correlation Latency Correlation between Test and Train devices} \\ \hline
{} & 1080ti\_64 & 2080ti\_1 & eyeriss & gold\_6226 & gold\_6240 & raspi4 & samsung\_s7 & silver\_4210r & titan\_rtx\_1 & titan\_rtx\_32 \\
\midrule
1080ti\_1        &      0.439 &     0.944 &   0.247 &      0.207 &      0.536 &  0.256 &       0.307 &         0.653 &         0.948 &          0.846 \\
pixel2           &      0.496 &     0.287 &   0.767 &      0.747 &      0.678 &  0.747 &       0.629 &         0.653 &         0.296 &          0.369 \\
essential\_ph\_1 &      0.414 &     0.240 &   0.678 &      0.670 &      0.663 &  0.660 &       0.650 &         0.608 &         0.268 &          0.313 \\
\bottomrule
\end{tabular}

}
\end{table*}

\begin{table*}[h!]
    \centering
    \resizebox{\linewidth}{!}{%
\begin{tabular}{cccccccccccccccc}\toprule
\multicolumn{16}{c}{FBNet FA Train-Test Correlation Latency Correlation between Test and Train devices} \\ \hline
{} & 1080ti\_1 & 1080ti\_32 & 1080ti\_64 & 2080ti\_1 & 2080ti\_32 & 2080ti\_64 & titan\_rtx\_1 & titan\_rtx\_32 & titan\_rtx\_64 & titanx\_1 & titanx\_32 & titanx\_64 & titanxp\_1 & titanxp\_32 & titanxp\_64 \\
\midrule
gold\_6226       &     0.207 &      0.241 &      0.323 &     0.178 &      0.214 &      0.297 &         0.184 &          0.209 &          0.274 &     0.232 &      0.270 &      0.344 &      0.222 &       0.250 &       0.303 \\
essential\_ph\_1 &     0.258 &      0.338 &      0.414 &     0.240 &      0.312 &      0.395 &         0.268 &          0.313 &          0.388 &     0.300 &      0.379 &      0.427 &      0.286 &       0.362 &       0.406 \\
samsung\_s7      &     0.307 &      0.372 &      0.421 &     0.293 &      0.349 &      0.406 &         0.312 &          0.352 &          0.404 &     0.347 &      0.402 &      0.435 &      0.337 &       0.388 &       0.416 \\
pixel2           &     0.300 &      0.409 &      0.496 &     0.287 &      0.388 &      0.485 &         0.296 &          0.369 &          0.466 &     0.328 &      0.449 &      0.512 &      0.318 &       0.425 &       0.486 \\
\bottomrule
\end{tabular}

}
\caption{FBNet task train-test correlations. Rows are test devices, Column headers are training devices.}
\label{tab:fbntraintestcorr}
\end{table*}

\begin{table*}[h!]
    \centering
    \resizebox{0.6\linewidth}{!}{%
            \begin{tabular}{@{}lccc@{}}
            \toprule
            Device & Type & NB201 & FBNet \\ 
            \midrule 
            \multicolumn{4}{c}{HELP \& HW-NAS-Bench~\cite{help, hwnasbench}}\\\midrule
            \texttt{1080ti\_1            }             & GPU &      \cmark & \cmark \\ 
            \texttt{2080ti\_1            }             & GPU &      \cmark & \cmark \\ 
            \texttt{1080ti\_32           }             & GPU &      \cmark & \cmark \\ 
            \texttt{2080ti\_32           }             & GPU &      \cmark & \cmark \\ 
            \texttt{1080ti\_256          }             & GPU &      \cmark & \cmark \\ 
            \texttt{2080ti\_256          }             & GPU &      \cmark & \cmark \\ 
            \texttt{titan\_rtx\_1        }             & GPU &      \cmark & \cmark \\ 
            \texttt{titanx\_1            }             & GPU &      \cmark & \cmark \\ 
            \texttt{titanxp\_1           }             & GPU &      \cmark & \cmark \\ 
            \texttt{titan\_rtx\_32       }             & GPU &      \cmark & \cmark \\ 
            \texttt{titanx\_32           }             & GPU &      \cmark & \cmark \\ 
            \texttt{titanxp\_32          }             & GPU &      \cmark & \cmark \\ 
            \texttt{titan\_rtx\_256      }             & GPU &      \cmark & \cmark \\ 
            \texttt{titanx\_256          }             & GPU &      \cmark & \cmark \\ 
            \texttt{titanxp\_256         }             & GPU &      \cmark & \cmark \\ \midrule
            \texttt{gold\_6240           }             & CPU &      \cmark & \cmark \\ 
            \texttt{silver\_4114         }             & CPU &      \cmark & \cmark \\ 
            \texttt{silver\_4210r        }             & CPU &      \cmark & \cmark \\ 
            \texttt{gold\_6226           }             & CPU &      \cmark & \cmark \\ \midrule
            \texttt{samsung\_a50         }             & mCPU &   \cmark & \cmark \\ 
            \texttt{pixel3               }             & mCPU &   \cmark & \cmark \\ 
            \texttt{samsung\_s7          }             & mCPU &   \cmark & \cmark \\ 
            \texttt{essential\_ph\_1     }             & mCPU &   \cmark & \cmark \\ 
            \texttt{pixel2               }             & mCPU &   \cmark & \cmark \\ \midrule
            \texttt{fpga                 }             & FPGA &     \cmark & \cmark \\ 
            \texttt{raspi4               }             & eCPU &    \cmark & \cmark \\ 
            \texttt{eyeriss              }             & ASIC &     \cmark & \cmark \\ \midrule
            \toprule
            Device & Type & NB201 & FBNet \\ 
            \midrule 
            \multicolumn{4}{c}{EAGLE\cite{brpnas}}\\\midrule
            \texttt{core\_i7\_7820x\_fp32}               & CPU  & \cmark & \xmark \\ \midrule
            \texttt{snapdragon\_675\_kryo\_460\_int8   } & mCPU   & \cmark & \xmark \\ 
            \texttt{snapdragon\_855\_kryo\_485\_int8   } & mCPU   & \cmark & \xmark \\ 
            \texttt{snapdragon\_450\_cortex\_a53\_int8 } & mCPU   & \cmark & \xmark \\ \midrule
            \texttt{edge\_tpu\_int8                    } & eTPU & \cmark & \xmark \\ \midrule
            \texttt{gtx\_1080ti\_fp32                  } & GPU  & \cmark & \xmark \\ \midrule
            \texttt{jetson\_nano\_fp16                 } & eGPU & \cmark & \xmark \\ 
            \texttt{jetson\_nano\_fp32                 } & eGPU & \cmark & \xmark \\ \midrule
            \texttt{snapdragon\_855\_adreno\_640\_int8 } & mGPU   & \cmark & \xmark \\ 
            \texttt{snapdragon\_450\_adreno\_506\_int8 } & mGPU   & \cmark & \xmark \\ 
            \texttt{snapdragon\_675\_adreno\_612\_int8 } & mGPU   & \cmark & \xmark \\ \midrule
            \texttt{snapdragon\_675\_hexagon\_685\_int8} & mDSP   & \cmark & \xmark \\ 
            \texttt{snapdragon\_855\_hexagon\_690\_int8} & mDSP   & \cmark & \xmark \\ 
            \bottomrule
            \end{tabular}%
}
\caption{Devices used in our paper and their categories (type). Note that we are referring to latency measurements on the same device with different batch sizes as a new device as well, this is because in some cases, there is a low correlation between these measurements.}
\label{tab:all_devicelist}
\end{table*}

\begin{table*}[h]
\centering
\begin{tabular}{ccl}
\toprule
\textbf{Task Index} & \textbf{Type} & \textbf{Devices} \\
\midrule
\multirow{15}{*}{ND} & Train & \texttt{1080ti\_1} \\
& & \texttt{1080ti\_32} \\
& & \texttt{1080ti\_256} \\
& & \texttt{silver\_4114} \\
& & \texttt{silver\_4210r} \\
& & \texttt{samsung\_a50} \\
& & \texttt{pixel3} \\
& & \texttt{essential\_ph\_1} \\
& & \texttt{samsung\_s7} \\ \cmidrule(lr){2-3} \addlinespace[0.5ex]
& Test & \texttt{titan\_rtx\_256} \\
& & \texttt{gold\_6226} \\
& & \texttt{fpga} \\
& & \texttt{pixel2} \\
& & \texttt{raspi4} \\
& & \texttt{eyeriss} \\ \midrule
\multirow{10}{*}{N1} & Train & \texttt{embedded\_tpu\_edge\_tpu\_int8} \\
& & \texttt{eyeriss} \\
& & \texttt{mobile\_gpu\_snapdragon\_675\_adreno\_612\_int8} \\
& & \texttt{mobile\_gpu\_snapdragon\_855\_adreno\_640\_int8} \\
& & \texttt{pixel3} \\ \cmidrule(lr){2-3} \addlinespace[0.5ex]
& Test & \texttt{1080ti\_1} \\
& & \texttt{titan\_rtx\_32} \\
& & \texttt{titanxp\_1} \\
& & \texttt{2080ti\_32} \\
& & \texttt{titan\_rtx\_1} \\ \midrule
\multirow{10}{*}{N2} & Train & \texttt{1080ti\_1} \\
& & \texttt{1080ti\_32} \\
& & \texttt{titanx\_32} \\
& & \texttt{titanxp\_1} \\
& & \texttt{titanxp\_32} \\ \cmidrule(lr){2-3} \addlinespace[0.5ex]
& Test & \texttt{embedded\_gpu\_jetson\_nano\_fp16} \\
& & \texttt{embedded\_tpu\_edge\_tpu\_int8} \\
& & \texttt{mobile\_dsp\_snapdragon\_675\_hexagon\_685\_int8} \\
& & \texttt{mobile\_dsp\_snapdragon\_855\_hexagon\_690\_int8} \\
& & \texttt{pixel3} \\ \midrule
\multirow{10}{*}{N3}  & Train & \texttt{desktop\_gpu\_gtx\_1080ti\_fp32} \\
& & \texttt{embedded\_gpu\_jetson\_nano\_fp16} \\
& & \texttt{eyeriss} \\
& & \texttt{mobile\_dsp\_snapdragon\_675\_hexagon\_685\_int8} \\
& & \texttt{mobile\_gpu\_snapdragon\_855\_adreno\_640\_int8} \\ \cmidrule(lr){2-3} \addlinespace[0.5ex]
& Test & \texttt{1080ti\_1} \\
& & \texttt{2080ti\_1} \\
& & \texttt{titanxp\_1} \\
& & \texttt{2080ti\_32} \\
& & \texttt{titanxp\_32} \\ \bottomrule
\end{tabular}
\caption{Hardware devices for NASBench-201}
\end{table*}

\begin{table*}[h]
\centering
\begin{tabular}{ccl}
\toprule
\textbf{Task Index} & \textbf{Type} & \textbf{Devices} \\
\midrule
\multirow{13}{*}{N4} & Train & \texttt{desktop\_cpu\_core\_i7\_7820x\_fp32} \\
& & \texttt{embedded\_gpu\_jetson\_nano\_fp32} \\
& & \texttt{embedded\_tpu\_edge\_tpu\_int8} \\
& & \texttt{eyeriss} \\
& & \texttt{mobile\_cpu\_snapdragon\_855\_kryo\_485\_int8} \\
& & \texttt{mobile\_dsp\_snapdragon\_675\_hexagon\_685\_int8} \\
& & \texttt{mobile\_dsp\_snapdragon\_855\_hexagon\_690\_int8} \\
& & \texttt{mobile\_gpu\_snapdragon\_675\_adreno\_612\_int8} \\
& & \texttt{mobile\_gpu\_snapdragon\_855\_adreno\_640\_int8} \\
& & \texttt{pixel2} \\ \cmidrule(lr){2-3} \addlinespace[0.5ex]
& Test & \texttt{1080ti\_1} \\
& & \texttt{2080ti\_1} \\
& & \texttt{titan\_rtx\_1} \\ \midrule
\multirow{20}{*}{N2}  & Train & \texttt{titan\_rtx\_1} \\
& & \texttt{titan\_rtx\_32} \\
& & \texttt{titanxp\_1} \\
& & \texttt{2080ti\_1} \\
& & \texttt{titanx\_1} \\
& & \texttt{1080ti\_1} \\
& & \texttt{titanx\_32} \\
& & \texttt{titanxp\_32} \\
& & \texttt{2080ti\_32} \\
& & \texttt{1080ti\_32} \\
& & \texttt{gold\_6226} \\
& & \texttt{samsung\_s7} \\
& & \texttt{silver\_4114} \\
& & \texttt{gold\_6240} \\
& & \texttt{silver\_4210r} \\
& & \texttt{samsung\_a50} \\
& & \texttt{pixel2} \\ \cmidrule(lr){2-3} \addlinespace[0.5ex]
& Test & \texttt{eyeriss} \\
& & \texttt{desktop\_gpu\_gtx\_1080ti\_fp32} \\
& & \texttt{embedded\_tpu\_edge\_tpu\_int8} \\ \bottomrule
\end{tabular}
\caption{Hardware devices for NASBench-201}
\end{table*}

\begin{table*}[h]
\centering
\caption{Hardware devices for FBNet}\vspace{2mm}
  \resizebox{0.7\linewidth}{!}{%
\begin{tabular}{ccl}
\toprule
\textbf{Task Index} & \textbf{Type} & \textbf{Devices} \\
\midrule
\multirow{12}{*}{FD} & Train & \texttt{1080ti\_1} \\
& & \texttt{1080ti\_32} \\
& & \texttt{1080ti\_64} \\
& & \texttt{silver\_4114} \\
& & \texttt{silver\_4210r} \\
& & \texttt{samsung\_a50} \\
& & \texttt{pixel3} \\
& & \texttt{essential\_ph\_1} \\
& & \texttt{samsung\_s7} \\ \cmidrule(lr){2-3} \addlinespace[0.5ex]
& Test & \texttt{fpga} \\
& & \texttt{raspi4} \\
& & \texttt{eyeriss} \\ \midrule
\multirow{10}{*}{F1} & Train & \texttt{2080ti\_1} \\
& & \texttt{essential\_ph\_1} \\
& & \texttt{silver\_4114} \\
& & \texttt{titan\_rtx\_1} \\
& & \texttt{titan\_rtx\_32} \\ \cmidrule(lr){2-3} \addlinespace[0.5ex]
& Test & \texttt{eyeriss} \\
& & \texttt{fpga} \\
& & \texttt{raspi4} \\
& & \texttt{samsung\_a50} \\
& & \texttt{samsung\_s7} \\ \midrule
\multirow{10}{*}{F2}  & Train & \texttt{essential\_ph\_1} \\
& & \texttt{gold\_6226} \\
& & \texttt{gold\_6240} \\
& & \texttt{pixel3} \\
& & \texttt{raspi4} \\ \cmidrule(lr){2-3} \addlinespace[0.5ex]
& Test & \texttt{1080ti\_1} \\
& & \texttt{1080ti\_32} \\
& & \texttt{2080ti\_32} \\
& & \texttt{titan\_rtx\_1} \\
& & \texttt{titanxp\_1} \\ \midrule
\multirow{10}{*}{F3} & Train & \texttt{essential\_ph\_1} \\
& & \texttt{pixel2} \\
& & \texttt{pixel3} \\
& & \texttt{raspi4} \\
& & \texttt{samsung\_s7} \\ \cmidrule(lr){2-3} \addlinespace[0.5ex]
& Test & \texttt{1080ti\_1} \\
& & \texttt{1080ti\_32} \\
& & \texttt{2080ti\_1} \\
& & \texttt{titan\_rtx\_1} \\
& & \texttt{titan\_rtx\_32} \\ \bottomrule
\end{tabular}
\centering
\begin{tabular}{ccl}
\toprule
\textbf{Task Index} & \textbf{Type} & \textbf{Devices} \\
\midrule
\multirow{13}{*}{F4}  & Train & \texttt{1080ti\_64} \\
& & \texttt{2080ti\_1} \\
& & \texttt{eyeriss} \\
& & \texttt{gold\_6226} \\
& & \texttt{gold\_6240} \\
& & \texttt{raspi4} \\
& & \texttt{samsung\_s7} \\
& & \texttt{silver\_4210r} \\
& & \texttt{titan\_rtx\_1} \\
& & \texttt{titan\_rtx\_32} \\ \cmidrule(lr){2-3} \addlinespace[0.5ex]
& Test & \texttt{1080ti\_1} \\
& & \texttt{pixel2} \\
& & \texttt{essential\_ph\_1} \\ \midrule
\multirow{19}{*}{FA}  & Train & \texttt{1080ti\_1} \\
& & \texttt{1080ti\_32} \\
& & \texttt{1080ti\_64} \\
& & \texttt{2080ti\_1} \\
& & \texttt{2080ti\_32} \\
& & \texttt{2080ti\_64} \\
& & \texttt{titan\_rtx\_1} \\
& & \texttt{titan\_rtx\_32} \\
& & \texttt{titan\_rtx\_64} \\
& & \texttt{titanx\_1} \\
& & \texttt{titanx\_32} \\
& & \texttt{titanx\_64} \\
& & \texttt{titanxp\_1} \\
& & \texttt{titanxp\_32} \\
& & \texttt{titanxp\_64} \\ \cmidrule(lr){2-3} \addlinespace[0.5ex]
& Test & \texttt{gold\_6226} \\
& & \texttt{essential\_ph\_1} \\
& & \texttt{samsung\_s7} \\
& & \texttt{pixel2} \\ \bottomrule
\end{tabular}%
}
\end{table*}

\end{document}